\documentclass{article}

\usepackage{microtype}
\usepackage{graphicx}
\usepackage{booktabs} %
\usepackage{subcaption}
\usepackage{hyperref}
\usepackage{amsmath}
\usepackage{amssymb}
\usepackage{mathtools}
\usepackage{amsthm}
\usepackage{multirow}
\usepackage{multicol}
\usepackage{makecell}
\usepackage{colortbl}
\usepackage{xcolor}
\usepackage{wrapfig}

\definecolor{darkblue}{rgb}{0.121, 0.316, 0.590}
\newcommand{\smallscript}[1]{{\scriptsize\textcolor{darkblue}{\textbf{#1}}}}

\renewcommand{\paragraph}[1]{\vspace{-2mm}\noindent\textbf{#1}}
\newlength\savewidth\newcommand\shline{\noalign{\global\savewidth\arrayrulewidth
  \global\arrayrulewidth 1pt}\hline\noalign{\global\arrayrulewidth\savewidth}}
\newcolumntype{x}[1]{>{\centering\arraybackslash}p{#1pt}}
\newcolumntype{y}[1]{>{\raggedright\arraybackslash}p{#1pt}}
\newcolumntype{z}[1]{>{\raggedleft\arraybackslash}p{#1pt}}
\definecolor{mygreen}{RGB}{0, 205, 108}
\definecolor{convcolor}{HTML}{412F8A}
\definecolor{resnetcolor}{HTML}{8DA0CB}
\definecolor{vitcolor}{HTML}{fc8e62}
\newcommand{\convcolor}[1]{\textcolor{convcolor}{#1}}
\newcommand{\cb}{\convcolor{$\bullet$\,}}

\definecolor{mygreen}{RGB}{0, 205, 108}
\definecolor{code0}{RGB}{223, 227, 241}
\definecolor{code1}{RGB}{250, 219, 223}
\definecolor{convcolor}{HTML}{412F8A}
\definecolor{resnetcolor}{HTML}{8DA0CB}
\definecolor{vitcolor}{HTML}{fc8e62}

\usepackage{pifont}       %
\usepackage{bbding}       %
\usepackage{fontawesome}
\usepackage{xspace}

\hypersetup{
    colorlinks=true,
    linkcolor=red,
    citecolor=blue,
    filecolor=magenta,      
    urlcolor=magenta,
    }
\newcommand{\tablestyle}[2]{\setlength{\tabcolsep}{#1}\renewcommand{\arraystretch}{#2}\centering\footnotesize}

\definecolor{orange}{HTML}{ff7f0e}
\definecolor{blue}{HTML}{1f77b4}
\definecolor{baselinecolor}{gray}{.9}
\newcommand{\baseline}[1]{\cellcolor{baselinecolor}{#1}}

\usepackage[accepted]{icml2025}

\usepackage{amsmath}
\usepackage{amssymb}
\usepackage{mathtools}
\usepackage{amsthm}

\usepackage[capitalize,noabbrev]{cleveref}

\theoremstyle{plain}

\theoremstyle{definition}

\theoremstyle{remark}

\hypersetup{
    colorlinks=true,
    linkcolor=red,
    citecolor=blue,
    filecolor=magenta,      
    urlcolor=magenta,
    }

\usepackage[textsize=tiny]{todonotes}

\begin{document}

\twocolumn[
\icmltitle{Scaling Up Parameter Generation: A Recurrent Diffusion Approach}

\begin{icmlauthorlist}
\icmlauthor{Kai Wang$^{1*}$}{}
\icmlauthor{Dongwen Tang$^{1*}$}{}
\icmlauthor{Wangbo Zhao$^{1}$}{}
\icmlauthor{Konstantin Schürholt$^{2}$}{}
\icmlauthor{Zhangyang Wang$^{3\dagger}$}{}
\icmlauthor{Yang You$^{1\dagger}$}{}
\end{icmlauthorlist}

\begin{center}
$^1$National University of Singapore \quad $^2$University of St.Gallen \quad $^3$University of Texas at Austin\\
\end{center}

\icmlkeywords{Machine Learning, ICML}

\vskip 0.3in
]

{
\renewcommand{\thefootnote}{}
\footnotetext[0]{
$^{*}$equal contribution, $^{\dagger}$corresponding author}
\footnotetext[1]{\ Code: \href{https://github.com/NUS-HPC-AI-Lab/Recurrent-Parameter-Generation}{NUS-HPC-AI-Lab/RPG}}
}

\begin{abstract}

Parameter generation has long struggled to match the scale of today’s large vision and language models, curbing its broader utility. In this paper, we introduce \textit{\textbf{R}ecurrent Diffusion for Large-Scale \textbf{P}arameter \textbf{G}eneration (\textbf{RPG})}, a novel framework that generates full neural network parameters—up to \textbf{hundreds of millions}—on a \textbf{single GPU}. Our approach first partitions a network’s parameters into non-overlapping `tokens', each corresponding to a distinct portion of the model. A recurrent mechanism then learns the inter-token relationships, producing `prototypes' which serve as conditions for a diffusion process that ultimately synthesizes the full parameters. Across a spectrum of architectures and tasks—including ResNets, ConvNeXts and ViTs on ImageNet-1K and COCO, and even LoRA-based LLMs—RPG achieves performance on par with fully trained networks while avoiding excessive memory overhead. Notably, it generalizes beyond its training set to generate valid parameters for previously unseen tasks, highlighting its flexibility in dynamic and open-ended scenarios. By overcoming the longstanding memory and scalability barriers, RPG serves as a critical advance in `\textbf{AI generating AI}', potentially enabling efficient weight generation at scales previously deemed infeasible.

\end{abstract}

\section{Introduction}

\begin{figure}[tp]
    \centering
    \includegraphics[width=0.48\textwidth]{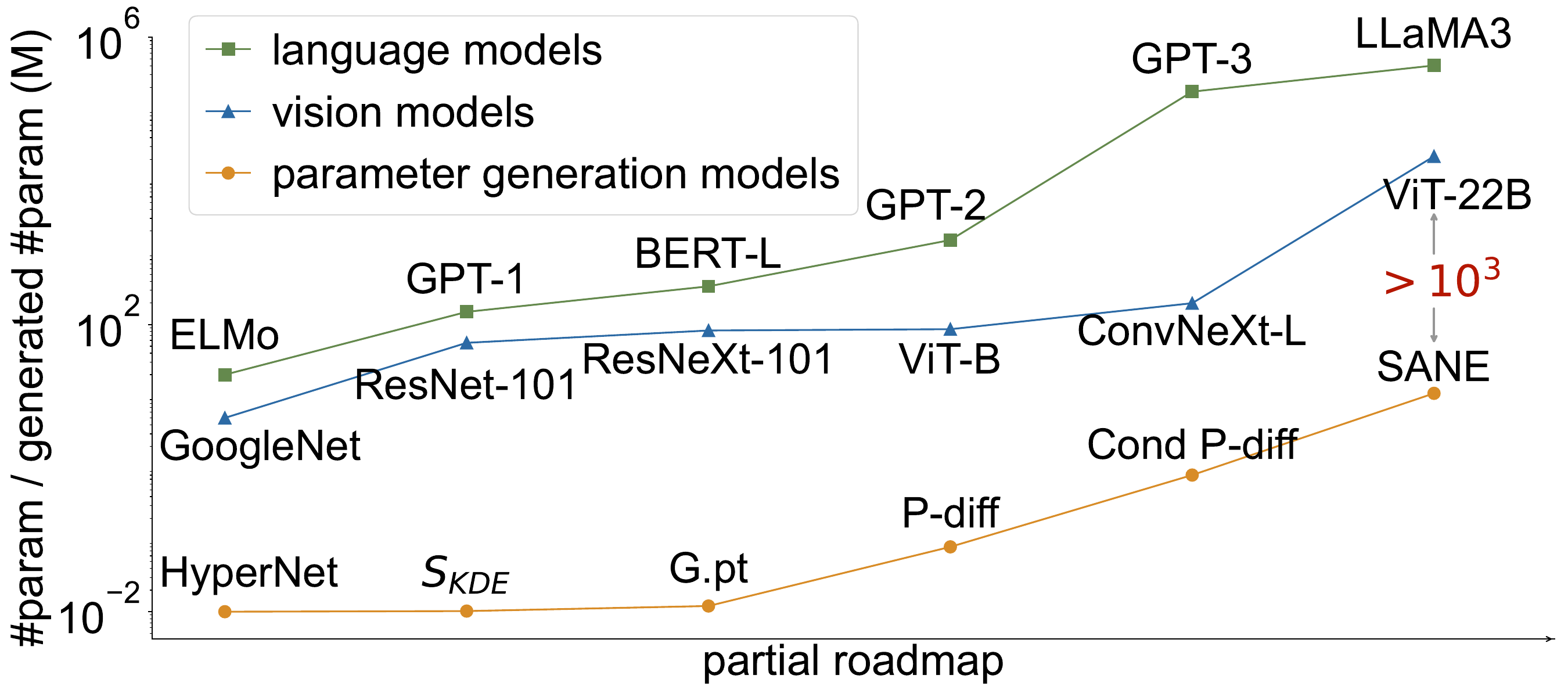}
    \vspace{-2em}
\caption{Partial roadmap of vision, language, and parameter generation models. Parameter number in vision or language models is at least $\mathbf{10^{3}}$ times larger than that of generated parameters.} 
\vspace{-1mm}
\label{fig:roadmap}
\end{figure}

Scaling up neural networks has been one of the most crucial factors driving the progress of deep learning, enabling remarkable performance across a wide range of tasks~\cite{krizhevsky2012imagenet, he2016deep, meta2024introducing}. 
However, \emph{neural network parameter generation}---from early HyperNetworks~\cite{ha2017hypernetworks} to more recent diffusion-based methods~\cite{peebles2022learning, wang2024neural, soro2024diffusion}---has not kept pace with the expansion of mainstream vision and language models. Indeed, as illustrated in Fig.~\ref{fig:roadmap}, there is an astonishing scale gap of at least $10^3$ between the parameter counts of widely used models (\textit{e.g.,} ViT, ConvNeXt, LLaMA) and the largest parameter generators available today. This discrepancy not only underscores the challenge of \emph{scalability} but also confines existing generators to academic demonstrations rather than broad, real-world applicability.

Traditional approaches such as HyperNetworks struggle to handle even moderately sized models, primarily due to memory overheads and optimization complexity. While recent diffusion-based works attempt to break the memory bottleneck via partial-parameter or two-stage generation~\cite{wang2024neural, soro2024diffusion}, such methods may overlook key correlations among different parameter subsets. 
To illustrate the importance of capturing inter-part correlations, we conduct a simple experiment: we train two ViT-Tiny~\cite{dosovitskiy2021an} models (A and B) on CIFAR-10~\cite{krizhevsky2009learning}, then exchange half of their layer parameters. 
Despite both models performing well originally (\textit{i.e.,} $90.4\%$ for model A and $89.6\%$ for model B), the hybrid model's accuracy drops dramatically to $45.8\%$ (as seen below).
\begin{center}\vspace{-0.8em}
\tablestyle{6pt}{1.2}
\begin{tabular}{lccc}
model & ViT-Tiny A & partial A + partial B & ViT-Tiny B \\
\shline
acc. (\%) & 90.4 & 45.8 & 89.6 \\
\end{tabular}\vspace{-0.5em}
\end{center}
This striking performance gap highlights that adequately modeling parameter correlations is not a trivial detail: a robust generator must therefore \emph{explicitly model} such relationships while also surmounting the memory pitfalls of large-scale processing.

\paragraph{Our contributions.} 
We present \textit{Recurrent Diffusion for Large-Scale Parameter Generation (RPG)}, a novel, end-to-end framework capable of generating \emph{full} neural network parameters with up to \textbf{hundreds of millions} of weights using just a \textbf{single commodity GPU}. 
To the best of our knowledge, RPG is \textbf{the first} method to efficiently synthesize the parameters of models like ConvNeXt-L and LLaMA-LoRA--with up to $\sim$\textbf{200M} weights--in a \textbf{single pass}. This significant leap is achieved by two core technical innovations:
\vspace{-5pt}
\begin{enumerate}
    \setlength\itemsep{3pt}
    \item \textbf{Parameter processing for large-scale tokenization.} 
    We devise a highly customized \emph{parameter tokenization} strategy designed to preserve both the \emph{layer-wise distribution} and the \emph{cross-layer correlations}. 
    Specifically, we (i) split weights according to their layer indices, (ii) apply layer-wise normalization to mitigate distribution shifts, (iii) slice parameters into non-overlapping \emph{tokens} with uniform size, and (iv) introduce a lightweight \emph{permutation state} to alleviate  symmetry issues when collecting multiple checkpoints. 
    Additionally, we employ \emph{2D position embeddings} (layer index + token index) to ensure the network retains positional awareness of each token within the entire set. 
    This tailored pre-processing pipeline is crucial for learning stable and high-quality parameter representations.
    \item \textbf{Recurrent diffusion modeling.}
    Unlike prior diffusion-based generators that flatten or divide all parameters into a single sequence or chunks, we propose a \emph{recurrent} mechanism to learn the relationships among tokens. 
    Concretely, we map each token into a hidden space via a \emph{recurrent model}, whose outputs serve as \emph{prototypes}. 
    These prototypes then condition a 1D \emph{diffusion process} that progressively denoises the parameter tokens. 
    By recurrently modeling global inter-token interactions and leveraging a diffusion sampler for synthesis, RPG aligns well with the dual goals of capturing token-to-token dependencies and remaining memory-efficient.
\end{enumerate}
\vspace{-8pt}
\paragraph{Key results and impact.}
Extensive evaluation on Image-Net classification, ADE20K segmentation, COCO detection, and commonsense reasoning  show that RPG consistently produces weights \emph{on par} with trained models. Notably:
\begin{itemize}
    \setlength\itemsep{2pt}
    \item \textbf{Single-GPU feasibility.} Our recurrent design and parameter-token strategy enable inference on a single commodity GPU beyond 100M weights.
    \item \textbf{Generalization to unseen tasks.} RPG can generate high-performing networks for \emph{unseen} binary classification tasks on CIFAR-10, providing evidence for its broader generalization capabilities.
    \item \textbf{Architectural Versatility.} Our method supports diverse families such as ResNet, ViT, ConvNeXt, and LoRA-based language models, making it a unified framework for parameter generation.
\end{itemize}
\vspace{-1em}
These results place us significantly closer to the vision of `\emph{AI creating AI}', where a single generative model can synthesize entire networks tailored for different tasks. We believe RPG’s strong scalability, lightweight memory footprint, and ability to accurately capture global parameter relationships may open new frontiers in fields like rapid architecture search, efficient model adaptation, and beyond.

\section{Method: Large-Scale Parameter Generation}
\label{sec:method}
\begin{figure*}[htp]
    \centering
    \includegraphics[width=0.73\textwidth]{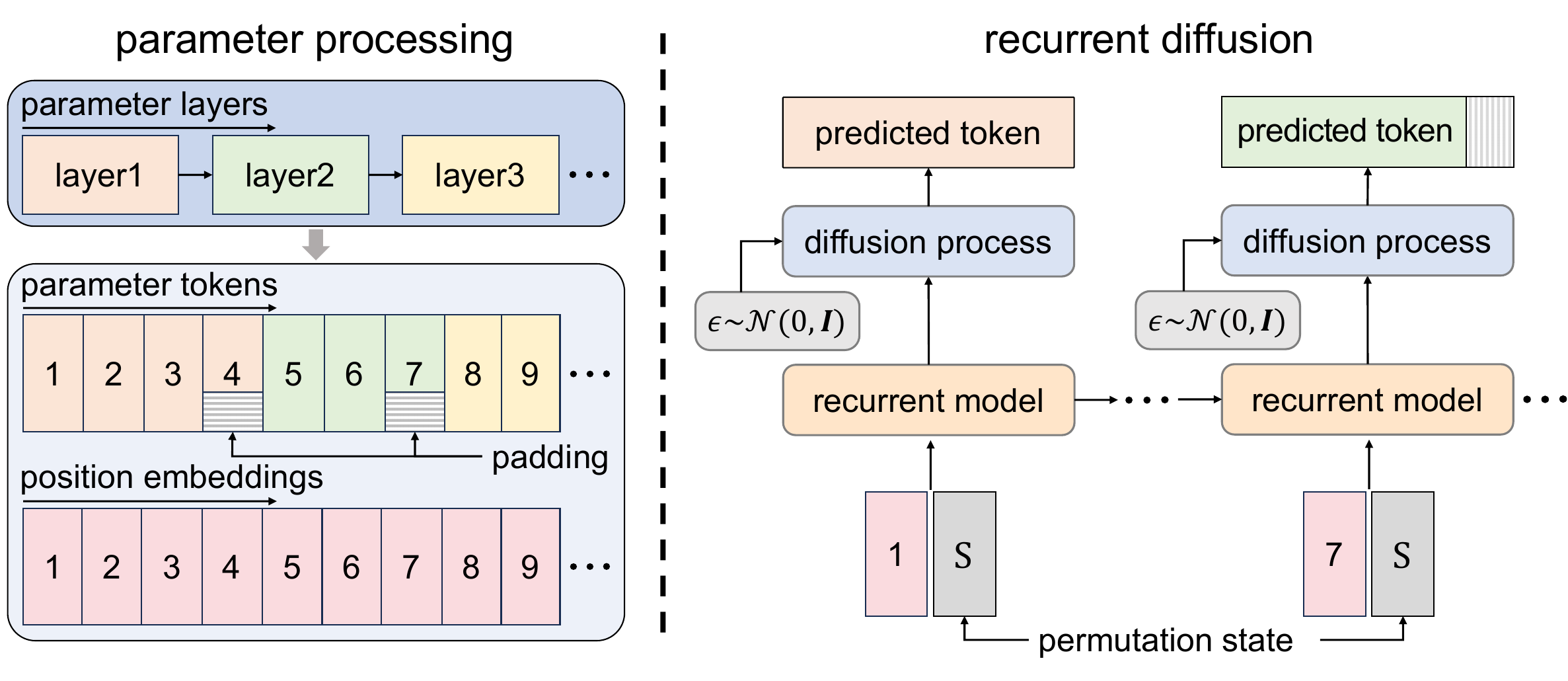}
    \vspace{-3mm}
\caption{Illustration of \textbf{parameter processing} (left) and inference of \textbf{recurrent diffusion} (right). The recurrent model integrates permutation states and position embeddings, generating prototypes that condition the diffusion model to synthesize the full parameters.}
\vspace{-2mm}
\label{fig:main}
\end{figure*}
In this section, we describe our proposed \emph{Recurrent Diffusion for Large-Scale Parameter Generation (RPG)}, which synthesizes \emph{full} neural network parameters under strict memory constraints. As illustrated in Fig.~\ref{fig:main}, our approach comprises two primary components: \textbf{parameter processing} and \textbf{recurrent diffusion}. 

\textbf{Parameter processing} (Sec.~\ref{re-diff:pt}) equips the model with a specialized tokenization pipeline, layer-wise normalization, permutation states, and position embeddings---all meticulously designed to mitigate issues like distribution shifts, neural symmetry, and the sheer size of modern network weights. 
\textbf{Recurrent diffusion} (Sec.~\ref{re-diff:prm}) then learns these `tokens' via a recurrent model whose outputs condition a 1D diffusion process, enabling us to capture \emph{global} parameter correlations while maintaining feasibility on a single GPU.

\subsection{Overview}
\vspace{-0.2em}
Our pipeline for large-scale parameter generation is shown conceptually on the right of Fig.~\ref{fig:main}.  
Given a set of trained parameters $\{W\}$ (e.g., from various checkpoints or model variants), we first transform them into \emph{parameter tokens}, augment them with a \emph{permutation state}, and apply \emph{position embeddings}. This processed sequence is fed into a \emph{recurrent model}, which outputs per-token `prototypes'. Finally, a \emph{diffusion model} uses these prototypes, along with random noise, to synthesize the entire parameter vector. In practice, we can sample from this generative model to produce a diverse family of high-performing weights.

\subsection{Parameter Processing}
\label{re-diff:pt}
\vspace{-0.5em}

\paragraph{Parameter tokenization.}
Drawing inspiration from the success of language and vision transformers~\cite{vaswani2017attention, dosovitskiy2021an}, we divide a network’s parameters into a sequence of non-overlapping \emph{tokens}. However, direct flattening or naive partitioning often fails to preserve crucial \emph{layer-wise} statistics and undermines global consistency. To circumvent this, we first segregate parameters by layer and then \emph{normalize} each layer to reduce distribution shifts:
\begin{equation}
    W \;\xrightarrow{\text{divide by layer}}\; [w[1], \dots, w[I]] 
    \;\xrightarrow{\text{$-\mu$, $/\sigma$}}\; [\hat{w}[1], \dots, \hat{w}[I]],
\end{equation}
where $w[i]$ and $\hat{w}[i]$ denote the original and normalized parameter blocks of the $i$-th layer, respectively, and $\mu,\sigma$ are the mean and standard deviation of that layer’s weights.

Next, each normalized layer $\hat{w}[i]$ is \emph{tokenized} into contiguous chunks of uniform size $k$ (plus padding if necessary):
\begin{equation}
    \hat{w}[i] \xrightarrow{\text{tokenize}} K[i] = \bigl[k_i^{1}, k_i^{2}, \dots, \mathrm{padding}\bigl(k_i^{J}\bigr)\bigr].
\end{equation}
Importantly, the padded regions are excluded from the loss and do not affect gradient updates. By enforcing a consistent token size across layers, we facilitate \emph{batch processing} of parameter tokens—a critical step for large-scale training.

\paragraph{Permutation state.}
When aggregating checkpoints from various training runs, \emph{neural symmetry}~\cite{kunin2021neural} can become a major bottleneck. Essentially, distinct permutations of layer weights may yield identical final performance but appear different to a generative model. To resolve this ambiguity, we introduce a distinct \emph{permutation state} $S$---encoded as a one-hot vector---for each checkpoint $W$. This extra condition guides the generative model to treat otherwise symmetrical weight configurations as unique exemplars, significantly stabilizing distribution learning.

\paragraph{Position embedding.}
To preserve token locality and inter-layer relationships, we augment each token with a \textbf{2D sinusoidal} position embedding~\cite{dosovitskiy2021an}. Specifically, for the $j$-th token of layer $i$, we assign
\begin{equation}
    K[i] \xrightarrow{\text{pos. embedding}} e[i] 
    = [\,e_i^{1}, \dots, e_i^{j}, \dots, e_i^{J}\,].
\end{equation}
The first dimension encodes the layer index, while the second dimension captures the token’s in-layer position. As a result, the g model knows not only how tokens are grouped by layer, but also each token’s placement within that layer.

\subsection{Recurrent Diffusion}
\label{re-diff:prm}
\vspace{-0.2em}

\paragraph{Recurrent model.}
After obtaining the parameter tokens $K[i]$, permutation states $S$, and position embeddings $e[i]$, we feed them into a \emph{recurrent} network $f(\cdot)$ that learns token-wise representations while capturing cross-token dependencies. We denote the output as a \emph{prototype} $P_i^j$:
\begin{equation}
    P_{i}^{j},\, H_{i}^{j} \;=\; f\!\Bigl(H_{i}^{j-1},\, e_{i}^{j},\, S\Bigr),
\end{equation}
where $H_{i}^{j}$ is the internal hidden state for the $j$-th token in layer $i$. In practice, we use Mamba~\cite{gu2024mamba} (a memory-efficient transformer alternative) followed by an MLP that projects features to the required dimension for our diffusion model. 
We find that a recurrence-based design elegantly handles long sequences of tokens, offering a strong balance between model capacity and memory usage. Later in experiments, we compare different recurrent architectures such as LSTM~\cite{LSTM} and causal-decoder transformers~\cite{vaswani2017attention}.

\paragraph{Parameter diffusion.}
Finally, we leverage a diffusion model to generate the actual parameter values. Drawing on p-diff~\cite{wang2024neural} and MAR~\cite{li2024autoregressive}, we construct a 1D convolutional network that accepts random noise $\epsilon$ together with the prototypes $P$. Formally, at diffusion time step $t$, we aim to predict the noise $\epsilon$ in a denoising objective:
\begin{equation}
    L_{\text{diff}} \;=\; \mathbb{E}_{t,\,K,\,\epsilon}\bigl\lVert\,\epsilon \;-\; 
    \epsilon_{\theta}\!\bigl(K_t,\; t,\; P\bigr)\bigr\rVert^2,
    \label{eq:diff_loss}
\end{equation}
where $\epsilon_\theta(\cdot)$ is our denoising network with learnable parameters $\theta$, and $K_t$ is the noisy version of $K$ at step $t$. Critically, the recurrent prototypes $P$ act as \emph{conditional inputs}, enforcing global coherence among tokens. Gradients from this diffusion loss flow back into the recurrent model, implicitly learning token correlations as well.

\paragraph{Inference.}
Once training is complete, the pipeline supports two inference modes. Basically, in conditional generation, we feed the permutation state $S$ of an existing checkpoint to replicate or perturb known parameter configurations. Further more, we can sample entirely \textbf{unseen task states} to produce novel parameter draws (Sec.~\ref{sec:4}). In both cases, the recurrent model outputs prototypes, and the diffusion model denoises random noise into a final parameter set that \emph{can scale up to hundreds of millions of weights}.

Overall, \textbf{recurrent diffusion} elegantly decouples the complex problem of \emph{global parameter correlation} into (i) a recurrent pass that focuses on cross-token interactions, and (ii) a diffusion pass that refines each token. This two-step design is precisely what allows RPG to handle large-scale architectures on standard hardware.

\section{Main Experiments}
\label{sec:experiments}

\subsection{Setup}

\paragraph{Datasets and architectures.}
We evaluate our method across various tasks, including ImageNet-1K~\cite{deng2009imagenet} for theclassification, ADE20K~\cite{zhou2017scene} for the semantic segmentation, COCO~\cite{lin2014microsoft} for the object detection, and BoolQ~\cite{clark2019boolq}, PIQA~\cite{bisk2020piqa}, SIQA~\cite{sap2019socialiqa}, HellaSwag~\cite{zellers2019hellaswag}, 
and ARC~\cite{clark2018think} for the commonsense reasoning tasks. 
To verify the scalability, we conduct experiments on various architectures with parameter counts ranging from several to hundred million.

\paragraph{Trained parameters collection.}
We take parameters collection on the ImageNet-1K as an example. 
To save the cost, we finetune the \textit{full parameters} of the models in timm~\footnote{https://github.com/huggingface/pytorch-image-models} and save 50 checkpoints as the training data. (More details are in Appendix~\ref{app_sec:data_collection})
For each checkpoint, we assign a unique permutation state to guide the generated parameters.

\begin{table}[t]
\centering
\tablestyle{3.5pt}{1.2}
\begin{tabular}{lccccccc}
arch.\textbackslash acc.(\%) &params. (M)& original & best & average & min \\
\shline
ResNet-18 &11.7&70.0 &69.9 &69.5 &69.0\\
ResNet-50 &25.6&79.8 &79.6 &79.5 &79.4 \\
ViT-Tiny &5.7 &74.9 &75.4 &75.3 &75.2 \\
ViT-Small &22.1 &81.4 &80.6 &80.5 &80.1 \\
ViT-Base &86.6 &84.4 &84.6 &84.4 &84.2 \\
ConvNeXt-Atto &3.7 &75.2 &74.6 &74.4 &74.2 \\
ConvNeXt-Large &197.8 &85.8 &85.8 &85.5 &85.2 \\

\end{tabular}
 \caption{We compare with the results of original models across seven architectures on the ImageNet-1K. Our approach successfully generates the entire model parameters that perform comparable results with the original models.}
\label{tab1}
\vspace*{-5mm}
\end{table}
\paragraph{Preprocessing and training details.}
The length of parameter tokens, permutation states, position embeddings, and prototypes is set to 8192.
It is worth noting that the permutation states and position embeddings are fixed during the training.
We also study the influence of the token length, varying it from 1024 to 16384.
The parameter diffusion consists of 1D convolutional layers. We default to using Mamba~\cite{gu2024mamba} as the architecture of the recurrent model. 
More details 
can be found in Appendix~\ref{app_sec:recipe}.%

\paragraph{Inference details.}
We input permutation states (relevant experimental results are in Appendix~\ref{app_sec:permutation_state}) and position embeddings into the recurrent model to generate the prototypes. Then, the diffusion model utilizes the prototypes as conditions, along with random noises, to synthesize the entire network parameters. We repeat the above process 10 times and report the best, average, and minimum.

\subsection{Results of Large-Scale Parameter Generation}

\paragraph{On ImageNet-1K.}
Tab.~\ref{tab1} presents performance comparisons
across seven architectures on ImageNet-1K. The parameter number of these architectures ranges from 3 to 197 million. 
Several observations can be made as follows: i) Our approach successfully generates model parameters at hundred-million scales, overcoming the out-of-memory issues faced by previous works~\cite{peebles2022learning, wang2024neural}.
ii) The performances of the generated models are comparable with the original ones.

\begin{table}[h]
\centering
\tablestyle{3.3pt}{1.2}
 \begin{tabular}{lcccc}

\multicolumn{1}{c}{\multirow{2}{*}{method}} &  \multicolumn{2}{c}{\multirow{1}{*}{ADE20K}} & \multicolumn{2}{c}{\multirow{1}{*}{COCO}}  \\ 

 \multicolumn{1}{c}{\multirow{1}{*}{}}  & \multicolumn{1}{c}{\multirow{1}{*}{mIoU(\%)}} & \multicolumn{1}{c}{\multirow{1}{*}{mAcc(\%)}}    & \multicolumn{1}{c}{\multirow{1}{*}{mAP Bbox (\%)}}  & \multicolumn{1}{c}{\multirow{1}{*}{mAP Seg (\%)}} \\ \shline

\multicolumn{1}{c}{\multirow{1}{*}{original}} &47.6 &58.3  &43.6 &39.0 \\
\multicolumn{1}{c}{\multirow{1}{*}{ours}}   &47.1 &57.5  &44.5 &39.6 \\
    
    \end{tabular}
\vspace{-1em}
\caption{Accuracy comparison of original and generated parameters on ADE20K (176.5M parameters) and COCO (110.9M parameters). All models are built based on ViT-Base.}
    \label{tab2}
\end{table}

\begin{table*}[h]
\centering
\tablestyle{6pt}{1.2}
\begin{tabular}{cc|cc|cc|cc|cc|cc|cc|cc}
\multirow{2}{*}{rank}& \multirow{2}{*}{params. (M)}& \multicolumn{2}{c|}{BoolQ} & \multicolumn{2}{c|}{PIQA} & \multicolumn{2}{c|}{SIQA} & \multicolumn{2}{c|}{HellaSwag} & \multicolumn{2}{c|}{ARC-e} & \multicolumn{2}{c|}{ARC-c} & \multicolumn{2}{c}{OBQA}\\
&& org & RPG 
& org & RPG 
& org & RPG
& org & RPG 
& org & RPG 
& org & RPG 
& org & RPG \\
\shline
4 &7.8 &\textbf{64.3} &63.1 &71.3 &\textbf{72.0} &66.0 &\textbf{67.5} &53.7 &\textbf{56.7} &64.4 &\textbf{65.3} &49.5 &\textbf{49.7} &63.1 &\textbf{66.0}\\
64 &113.1 &\textbf{69.3} &69.1 &78.9 &\textbf{79.4} &72.9 &\textbf{73.9} &81.1 &81.1 &72.9 &\textbf{73.1} &58.1 &\textbf{58.3} &71.2 &\textbf{72.1}\\
\end{tabular}
\caption{Accuracy comparisons of original and generated DoRA with varying ranks for LLaMA-7B on the commonsense reasoning tasks.
Our approach can generate comparable or even better results than original models.
\textbf{Bold entries} are best results.}
\label{commonsense reasoning}
\end{table*}

\paragraph{On ADE20K and COCO.}
For semantic segmentation, following ~\citet{zhao2024dynamic}, we adopt UperNet~\cite{xiao2018unified} as the segmentation model and train it on ADE20K~\cite{zhou2017scene} to prepare checkpoints. For object detection and instance segmentation, we finetune ViTDet~\cite{li2022exploring} on COCO~\cite{lin2014microsoft} to collect checkpoints and report the results of mAP Bbox and mAP Seg, respectively.  All experiments here are conducted based on ViT-Base~\cite{dosovitskiy2021an}. Tab.~\ref{tab2} presents the strong generalization of RPG to these two tasks. Specifically, compared to the original models, we achieve comparable or even slightly better results over all the above metrics.

\begin{table*}[h]
\centering
\hspace{-0.8em}
\subfloat[Recurrent model can largely improve the performance and stability. \label{tab:rm_ablation}]{
\centering
\begin{minipage}[h]{0.30\textwidth}
\begin{center}
\centering
\tablestyle{5pt}{1.15}

\begin{tabular}{lccc}%
     & best & avg. &min. \\
    \shline
    original  & 75.2 & 74.9 &  74.7 \\
     - recurrent model & fail & fail & fail \\
    \baseline{+ recurrent model}  & \baseline{\textbf{75.4}} & \baseline{\textbf{75.3}} & \baseline{\textbf{75.2}} \\
\end{tabular}
\end{center}
\end{minipage}
}
\hspace{1em}
\subfloat[Learnable embeddings perform better, but saving cost needs to be considered. \label{tab:pe_ablation}]{
\centering
\begin{minipage}[h]{0.30\textwidth}
\begin{center}
\tablestyle{4pt}{1.15}
\begin{tabular}{lccc}%
     pos. emb.& best & avg. &min. \\
    \shline
     learnable & \textbf{75.5} & \textbf{75.4} & \textbf{75.3} \\
     by index  & 75.4 & 75.3 &  75.0 \\
    \baseline{by layer} & \baseline{75.4} & \baseline{75.3} & \baseline{75.2} \\
\end{tabular}
\end{center}
\end{minipage}}
\hspace{1em}
\subfloat[Our tokenization achieves the best trade-off between performance and training time.\label{fig:manner of tokenization}]{
\centering
\begin{minipage}[h]{0.30\textwidth}
    \begin{center}
        \tablestyle{3.5pt}{1.15}
\begin{tabular}{lcccc}%
     tokenization & best & avg. &min. &time (h)  \\
     \shline
     flatten & 75.3 & 75.2 & 74.8 & 6.2\\
     by channel & 75.3 & 75.1 & 75.0 & 14.2\\
    \baseline{within layer} & \baseline{\textbf{75.4}} & \baseline{\textbf{75.3}} & \baseline{\textbf{75.2}} & \baseline{6.2}\\
\end{tabular}
    \end{center}
\end{minipage}
}
\vspace{-0.5em}
\caption{Ablation experiments of the recurrent model, position embeddings, and tokenization with ViT-Tiny on ImageNet-1K. Defaults are marked in \baseline{gray}.
`fail' indicates models performing at random-chance level.
\textbf{Bold entries} are best results.}
\label{tab:4_ablation}
\end{table*}

\paragraph{On commonsense reasoning.}
We employ DoRA~\cite{liu2024dora}, an upgraded version of LoRA~\cite{hu2022lora}, to fine-tune LLaMA-7B~\cite{touvron2023llama} for commonsense reasoning
tasks and save the checkpoints as the training data. 
We report the results across 7 sub-tasks with rank = 4 and 64 in Tab.~\ref{commonsense reasoning}. The generated models consistently yield results comparable to those of the original ones.

\subsection{Ablation Studies and Analysis}
In this section, 
we present the results of the generated ViT-Tiny on the ImageNet-1K, unless stated otherwise.

\paragraph{The effect of recurrent model.}
We employ the recurrent model to learn the relationship among parameter tokens. To keep other factors consistent, we simply remove the state transition function from the recurrent model for comparison, denoted as `$-$ recurrent model'. The experimental results from Tab.~\ref{tab:rm_ablation} confirm that the recurrent model plays a crucial role in parameter generation. Without the state transition function, our approach learns each parameter token individually, overlooking the relationships among these tokens. As a result, the generated parameters perform extremely poorly.

\paragraph{The manner of position embeddings.}
In ViT, the position embeddings are learnable by default. Here, we mainly conduct the experiments with three different position embedding manners and show the details as follows:

\cb learnable: Initializing with 2D sinusoidal positional encoding and set to be learnable.

\cb encoded by index: Using 1D sinusoidal positional encoding, irrespective of the original network structure, with indices assigned from front to back.

\cb encoded by layer (default): Using 2D sinusoidal positional encoding to represent layer and token indices.

As shown in Tab.~\ref{tab:pe_ablation}, the learnable embeddings perform slightly better than the other two manners. However, we still recommend using fixed position embeddings, as they offer comparable performance while significantly reducing storage requirements compared to the learnable one.

\paragraph{The manner of tokenization.}
Considering the differences among various layers, 
we divide the parameters into tokens within each layer. 
P-diff~\cite{wang2024neural} flattens the parameters into 1D vectors, while SANE~\cite{schurholt2024towards} divides the parameters by channel within each layer.
We compare the results of these 3 strategies in Tab.~\ref{fig:manner of tokenization}. Our default strategy achieves better results than the others. Flattening results in a single token containing parameters from different layers, which poses challenges for optimization. Tokenizing by the channel may result in excessive padding values for each token, as the channel number is usually much smaller than the token size.

\paragraph{The structure of recurrent model.}
We mainly explore three structures of the recurrent model, including LSTM~\cite{LSTM}, Transformer~\cite{vaswani2017attention}, and Mamba~\cite{gu2024mamba}. In Tab.~\ref{recurrent_structure}, we report the performances, training time, and memory cost of generating ViT-Tiny parameters on ImageNet-1K. 
All three structures can achieve good results. However, considering the training time and memory cost, our default Mamba is the best structure of the recurrent model.

\begin{table}[h]
    \centering

   \vspace{0mm}
\tablestyle{5pt}{1.2}
  \begin{tabular}{l| c c  c |c c}
    structure & best &avg. &min. & time (h) &memory (GB)\\
\shline
    LSTM%
    & \textbf{75.5} & 75.2 & 74.4 & 16.1  & 38.1 \\
    Transformer%
    & 75.0 & 74.8 & 74.6 & 4.2 & 29.1 \\
    \baseline{Mamba}%
    & \baseline{75.4} & \baseline{\textbf{75.3}} & \baseline{\textbf{75.2}} & \baseline{\textbf{4.1}} & \baseline{\textbf{27.8}} \\

    \end{tabular}
     \caption{We study the characteristics of three recurrent structures. Defaults are marked in \baseline{gray}.
\textbf{Bold entries} are best results.}
    \label{recurrent_structure}
\end{table}

\paragraph{Token size. }%
In Tab.~\ref{tab token size}, we show the results of generated ViT series with tokens of different sizes ranging from 1024 to 16384. 
The performance of generated models become better as the token size increases. When token is of small size, itcontains limited information that is hard to learn. 
Despite carrying large volume of information, larger token leads to substantial memory costs (see in Appendix Fig.~\ref{appendix_token_size_memory}). 

\begin{table}[h] %
  \centering %
\begin{minipage}{1.0\linewidth}
  \centering

\tablestyle{4.0pt}{1.2}
\vspace{0mm}
\begin{tabular}{cc| ccccc}
\multirow{2}{*}{model} &\multirow{2}{*}{params. (M)}  &\multicolumn{5}{c}{\multirow{1}{*}{token size}}\\

&& 1024 & 2048 & 4096 & 8192 & 16384 \\
\shline
ViT-Tiny &5.7  & 0.3 & 70.8 & 75.2 & \baseline{\textbf{75.3}} & 69.3 \\
ViT-Small &22.1  & 0.1 & 0.7 & 80.5 & \baseline{}{80.5} & 80.4 \\
ViT-Base &86.6 & 0.1 & 0.1 & 0.2 & 45.3 & \baseline{\textbf{84.4}} \\
\end{tabular}
\caption{
 Accuracy of generated models with the different toke sizes.
 Large token size performs better on large models. \textbf{Bold entries} are best results.}
\label{tab token size}
\end{minipage}

\end{table}

\paragraph{Efficiency of generating large-scale parameters.}
Rapid synthesis of large-scale parameters is crucial for evaluating the practicality of our approach. 
As illustrated in Tab.~\ref{tab: result_steps}, we present the time cost for generating models of ViT-Base and ConvNeXt-L across various DDIM~\cite{song2020denoising} sampling steps.
All results are obtained with a single NVIDIA H100 80G GPU.
Our approach shows the capability to generate models within minutes. Notably, even for ConvNeXt-L (197.7 M parameters), we can synthesize the entire parameter within 1.3 minutes.
Even with only 20 sampling steps, we can achieve promising results. Meanwhile, the inference memory requirement is approximately 20GB, so RPG can be deployed on NVIDIA RTX 3090 or similar-level GPUs.

\begin{table}[h]
    \centering

\begin{center}
\vspace{0em}
\tablestyle{2.5pt}{1.2}
\begin{tabular}{l| cccc|  cccc}
model / memory & \multicolumn{4}{c|}{ViT-Base / 20.7GB}  & \multicolumn{4}{c}{ConvNeXt-L / 21.6GB} \\
\shline
diffusion steps  & 20 & 60 & 100 & 200   & 20 & 60 & 100 & 200\\
time (minute)  & 0.6 & 0.8 & 1.1 & 1.8  & 0.7 & 1.3 & 2.0 & 3.5 \\
accuracy (\%)  & 83.3 & 84.4 & 84.4 & 84.3 & 85.0 & 85.5 & 85.3 & 85.3

\end{tabular}

\caption{GPU memory and inference time comparisons among different diffusion steps. Our approach can efficiently generate the entire ConvNeXt-L parameters (197.8 M) in minutes, requiring only $\sim$20GB of GPU memory.}
\label{tab: result_steps}
\end{center}
\end{table}

\paragraph{Similarity analysis of generated models.}
We compare RPG with adding noise and model interpolation.
Following p-diff~\citep{wang2024neural}, we use IoU to measure model similarity by comparing outputs across samples, selecting the maximum IoU with original models as our metric.

\begin{figure}[tp]
    \centering
    \begin{subfigure}{0.48\textwidth}
        \centering
        \includegraphics[width=\textwidth]{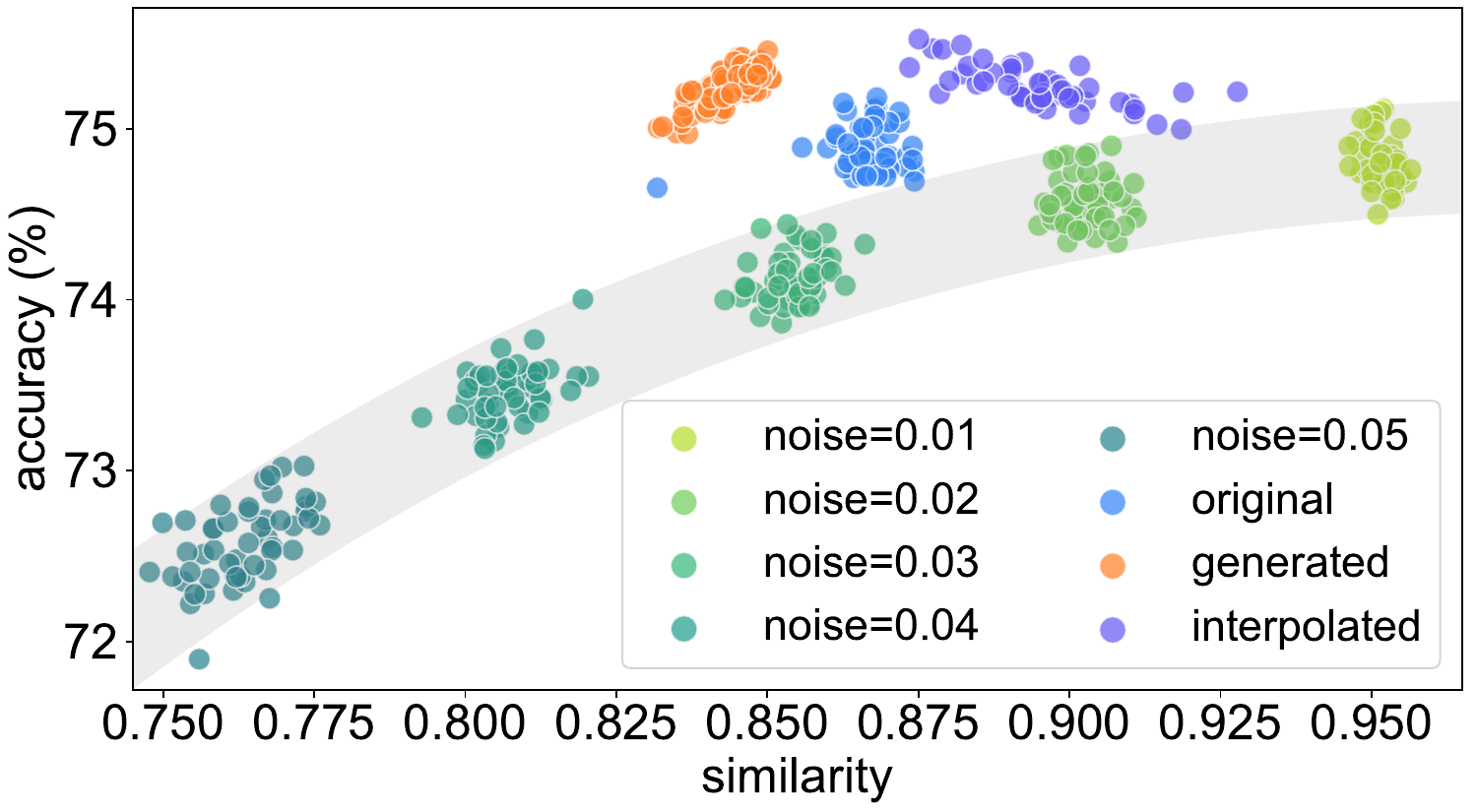}
        \label{fig:accuracy_similarity_vit_tiny}
    \end{subfigure}
    \vspace{-2em}
    \vspace{-1.5em}
    \caption{The figure shows the trade-off between accuracy and similarity with ViT-Tiny on ImageNet-1K. The \colorbox{gray!20}{shaded area} includes the approximate range of noise-added checkpoints. This plot demonstrates the strong trade-off between accuracy and similarity and highlights our advantages over trivial interpolation.}
    \label{fig:accuracy_similarity}
\end{figure}

In Fig.~\ref{fig:accuracy_similarity}, 
as the noise level increases, both the similarity and accuracy of the models decrease. The points representing our generated models are distributed in the \textit{upper left} region relative to the other conditions, indicating that our models can enhance diversity while maintaining accuracy. Furthermore, our method demonstrates greater diversity in comparison to trivial interpolation methods.

\subsection{Comparisons with Previous Methods}
We compare RPG with four previous works, \textit{i.e.}, $S_{\text{KDE30}}$~\cite{schurholt2022hyper}, p-diff~\cite{wang2024neural}, D2NWG~\cite{soro2024diffusion}, and SANE~\cite{schurholt2024towards}.
We report the original and generated performances for comparison.
As shown in Tab.~\ref{result_comparisons},
among all compared methods,
only our approach consistently achieves the highest results and comparable results to original models across various architectures. 
Another key issue is that the previous works usually fail to generate large-scale neural network parameters. (More discussion is in Appendix~\ref{app_sec:related}.)

\begin{table}[h]
    \centering

    \tablestyle{4.5pt}{1.2}
    \begin{tabular}{lc c c c}
    method &CNN (s) &CNN (m) & ResNet-18 & ViT-Base\\
    \shline
    params. (M) &0.003 &0.011 &11.7 &86.6\\
    \hline
    $S_{\text{KDE30}}$%
    &26.9 \smallscript{46.1} &- & OOM & OOM \\
    p-diff%
    &\underline{48.8} \smallscript{49.0} &\underline{61.9} \smallscript{62.1} & OOM & OOM \\
    SANE%
    & - &57.9 \smallscript{57.2} &68.6 \smallscript{85.5} & - \\
    D2NWG%
    & 38.2 \smallscript{44.7} & 58.8 \smallscript{57.2} &94.6 \smallscript{94.6}  & - \\
    RPG & 49.0 \smallscript{49.0} &62.0 \smallscript{62.1} &95.1 \smallscript{95.3} &98.9 \smallscript{98.7} \\
    \end{tabular}
    \vspace{-0.5em}
    \caption{
    Our approach obtains the best results across all architectures and largely outperforms most previous works. The \smallscript{subscripts} denote the average accuracy of corresponding original models. \underline{Underline} denotes the reproduced results. And OOM denotes out-of-memory issue. The detailed structures of CNN (s) and CNN (m) can be found in Model Zoos~\cite{Schurholt2022modelzoos}.}
    \label{result_comparisons}
\end{table}

\begin{figure*}[h]
    \centering
    \includegraphics[width=0.72\textwidth]{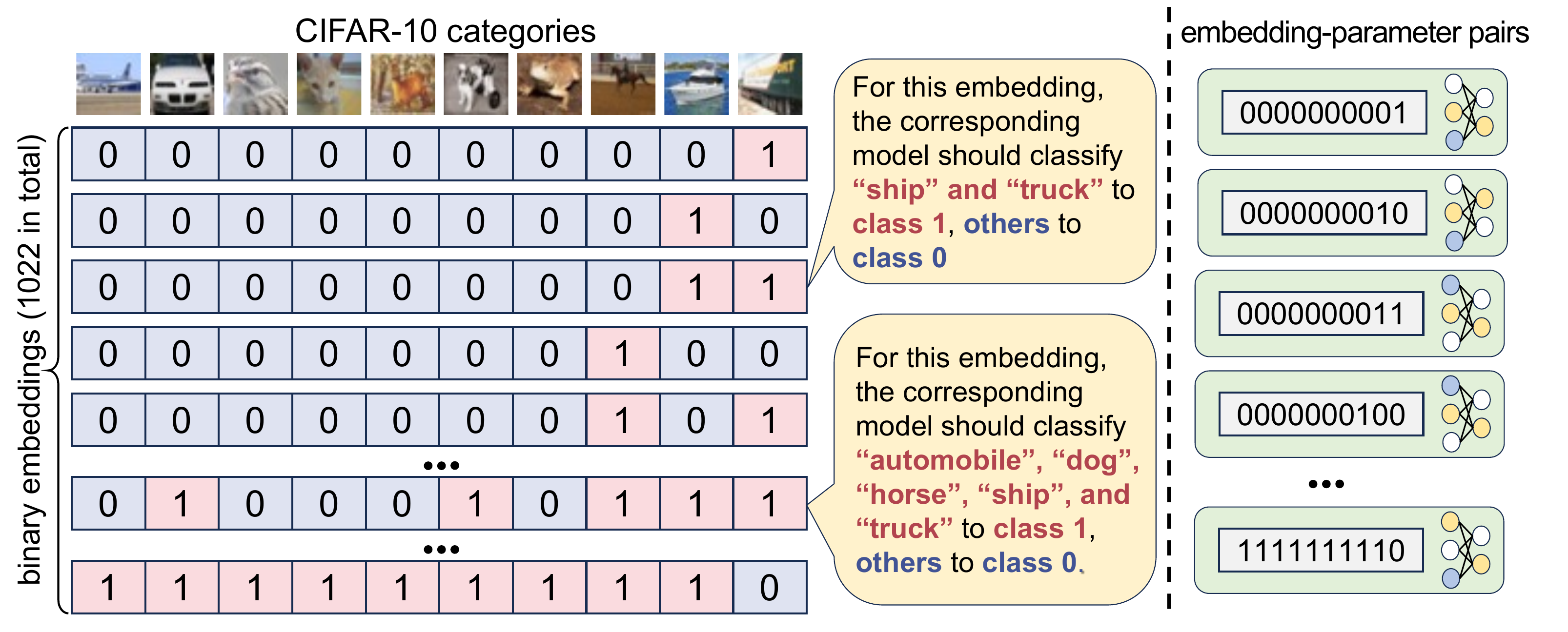}
    \caption{An illustration of our binary embedding strategy and dataset construction. Left: binary embeddings (1022 in total) encode different CIFAR-10 classification tasks, where 1s indicate classes to be classified together (e.g., `ship' and `truck' in the first example). Right: the dataset consists of parameter-encoding pairs, formed by network parameters with their corresponding binary embeddings. These pairs are split into non-overlapping training and validation sets.}
    \label{fig:divide_seen_and_unseen}
\end{figure*}

\section{RPG Generalizes to Unseen Tasks} 
\label{sec:4}
Until now, experimental results have demonstrated that our approach can efficiently generate large-scale neural network parameters if these models are included in the training set. In this section, we mainly investigate whether our approach has the ability to generate models to tackle unseen tasks.
This generalization capability is crucial as it enables RPG to handle novel tasks without requiring additional training, making it particularly valuable for real-world applications.

\subsection{Experiment Designs.}

\paragraph{Build seen and unseen tasks.} To assess RPG's capability 
for unseen tasks, we construct various binary classification tasks on CIFAR-10. As shown in Fig. ~\ref{fig:divide_seen_and_unseen}, we encode each task as a 10-bit binary embedding, where each bit corresponds to a CIFAR-10 category. 1 indicates that the corresponding category should be classified as positive (class 1), while 0 indicates negative (class 0). For example, in the third embedding shown, `ship' and `truck' are assigned to class 1, while all other categories belong to class 0.

Given this encoding strategy, we can create $2^{10}$ possible binary embeddings. After removing the two trivial cases (all 0s and all 1s), we obtain 1022 valid embeddings. For each embedding, we collect its corresponding model parameters, forming embedding-parameter pairs. These pairs are then split into \textit{non-overlapping} sets for training and validation, allowing us to evaluate RPG's generality on unseen tasks.

\paragraph{Collection of the checkpoints.}
We use ViT-Tiny to train 1022 binary classifiers on CIFAR-10 with different binary embeddings and save 3 models for each classifier. These binary embeddings serve as conditioning inputs for the subsequent RPG training process. Of these tasks, 1002 randomly selected embedding-parameter pairs serve as the training set (seen tasks), while the remaining pairs are reserved as unseen tasks for evaluation.

\paragraph{Training and evaluation of RPG.}
We only use the 1002 checkpoints from seen tasks to train RPG. Meanwhile, these embeddings are also fed into RPG as conditional inputs of the recurrent model. During the training stage, the checkpoints trained by the unseen binary embeddings are not accessible.
When evaluating, we input the unseen binary embeddings to the trained RPG to generate the parameters for unseen tasks.
The results of the original and generated unseen models are reported for comparison.  

\subsection{Results for Unseen Tasks}

\begin{table}[h]
    \centering
    \tablestyle{12pt}{1.2}
    \begin{tabular}{ccc}
        unseen tasks (embeddings)& original & RPG \\
        \hline
        \colorbox{code0!100}{0}\colorbox{code1!100}{1}\colorbox{code0!100}{0}\colorbox{code0!100}{0}\colorbox{code0!100}{0}\colorbox{code1!100}{1}\colorbox{code0!100}{0}\colorbox{code1!100}{1}\colorbox{code1!100}{1}\colorbox{code1!100}{1} & 97.3 & 94.4 \\
        \colorbox{code0!100}{0}\colorbox{code1!100}{1}\colorbox{code1!100}{1}\colorbox{code1!100}{1}\colorbox{code1!100}{1}\colorbox{code1!100}{1}\colorbox{code0!100}{0}\colorbox{code1!100}{1}\colorbox{code1!100}{1}\colorbox{code0!100}{0} & 98.1 & 96.6 \\
        \colorbox{code0!100}{0}\colorbox{code0!100}{0}\colorbox{code1!100}{1}\colorbox{code1!100}{1}\colorbox{code1!100}{1}\colorbox{code0!100}{0}\colorbox{code1!100}{1}\colorbox{code1!100}{1}\colorbox{code1!100}{1}\colorbox{code0!100}{0} & 97.4 & 95.0 \\ 
        \colorbox{code0!100}{0}\colorbox{code1!100}{1}\colorbox{code0!100}{0}\colorbox{code1!100}{1}\colorbox{code1!100}{1}\colorbox{code1!100}{1}\colorbox{code1!100}{1}\colorbox{code1!100}{1}\colorbox{code1!100}{1}\colorbox{code1!100}{1} & 98.4 & 96.1 \\ 
        \colorbox{code0!100}{0}\colorbox{code0!100}{0}\colorbox{code1!100}{1}\colorbox{code0!100}{0}\colorbox{code0!100}{0}\colorbox{code0!100}{0}\colorbox{code0!100}{0}\colorbox{code0!100}{0}\colorbox{code0!100}{0}\colorbox{code0!100}{0} & 98.9 & 96.6 \\
        \colorbox{code0!100}{0}\colorbox{code0!100}{0}\colorbox{code0!100}{0}\colorbox{code1!100}{1}\colorbox{code1!100}{1}\colorbox{code0!100}{0}\colorbox{code0!100}{0}\colorbox{code1!100}{1}\colorbox{code0!100}{0}\colorbox{code1!100}{1} & 96.7 & 92.9 \\
        \colorbox{code1!100}{1}\colorbox{code1!100}{1}\colorbox{code1!100}{1}\colorbox{code1!100}{1}\colorbox{code1!100}{1}\colorbox{code0!100}{0}\colorbox{code1!100}{1}\colorbox{code0!100}{0}\colorbox{code0!100}{0}\colorbox{code1!100}{1} & 97.6 & 94.8 \\ 
        \colorbox{code1!100}{1}\colorbox{code0!100}{0}\colorbox{code1!100}{1}\colorbox{code0!100}{0}\colorbox{code0!100}{0}\colorbox{code0!100}{0}\colorbox{code0!100}{0}\colorbox{code0!100}{0}\colorbox{code1!100}{1}\colorbox{code1!100}{1} & 98.1 & 95.7 \\ 
        \colorbox{code0!100}{0}\colorbox{code1!100}{1}\colorbox{code0!100}{0}\colorbox{code0!100}{0}\colorbox{code0!100}{0}\colorbox{code1!100}{1}\colorbox{code0!100}{0}\colorbox{code1!100}{1}\colorbox{code1!100}{1}\colorbox{code0!100}{0} & 97.1 & 93.6 \\ 
        \colorbox{code1!100}{1}\colorbox{code1!100}{1}\colorbox{code0!100}{0}\colorbox{code0!100}{0}\colorbox{code0!100}{0}\colorbox{code1!100}{1}\colorbox{code1!100}{1}\colorbox{code0!100}{0}\colorbox{code0!100}{0}\colorbox{code1!100}{1} & 97.0 & 94.0 \\
    \end{tabular}
    \caption{Result comparisons between original and generated models on unseen embeddings (tasks). Even without seeing trained parameters of unseen tasks, RPG directly generates models that achieve competitive performance on them.}
    \label{tab:unseen_comparison}
\end{table}
\paragraph{Performance comparisons.}
We compare the results of our approach and original models on unseen binary embeddings in Tab.~\ref{tab:unseen_comparison}. Considering the space limitation, we randomly select 10 unseen binary embeddings for comparison. Notably, RPG yields commendable performance in these unseen tasks, even without being trained on the specific unseen embeddings. That demonstrates the strong practicality and potential of our approach in generating models under unseen tasks. The results of the remaining unseen binary embeddings and more analysis are shown in Appendix~\ref{more_results_of_sec4}.

\begin{table*}[h]
    \centering
    \tablestyle{9pt}{1.2}
 \begin{tabular}{c c c  c c cccccc}

    binary embedding (from seen set)&\colorbox{code1!100}{1} &\colorbox{code0!100}{0} &\colorbox{code1!100}{1}  &\colorbox{code1!100}{1} &\colorbox{code1!100}{1} &\colorbox{code0!100}{0} &\colorbox{code1!100}{1} &\colorbox{code0!100}{0} &\colorbox{code0!100}{0} &\colorbox{code0!100}{0}\\

    accuracy (\%) &98.0 &99.1  &98.5 &94.4 &98.1 &92.2 &99.2 &97.0 &97.1 &99.1 \\
    \shline
    flipped embedding (from unseen set)&\colorbox{code0!100}{0} &\colorbox{code1!100}{1} &\colorbox{code0!100}{0} &\colorbox{code0!100}{0} &\colorbox{code0!100}{0}
    &\colorbox{code1!100}{1} &\colorbox{code0!100}{0}
    &\colorbox{code1!100}{1} &\colorbox{code1!100}{1} &\colorbox{code1!100}{1}\\

    \textcolor{black}{accuracy (\%)} &\textcolor{black}{92.3} &\textcolor{black}{98.9}  &\textcolor{black}{94.0} &\textcolor{black}{85.4} &\textcolor{black}{92.2} &\textcolor{black}{90.8} &\textcolor{black}{99.3} &\textcolor{black}{95.3} &\textcolor{black}{97.6} &\textcolor{black}{98.1} \\
    
    \end{tabular}
    \caption{Comparisons of binary embedding changes. The accuracy is for each single class. RPG can be aware of such changes accurately.}
    \label{perception_condition_changes}
\end{table*}

\paragraph{Perception of embedding changes.}
In addition to comparing results, we further investigate our approach's ability to perceive embedding changes. We select two tasks with opposite binary embeddings in each element and report the results in Tab.~\ref{perception_condition_changes}. Our approach demonstrates a remarkable capacity to accurately detect changes in the tasks and generate corresponding model parameters. It is worth noting that the accuracy would hover around 50\% if our approach were not aware of the embedding changes.

\begin{figure*}[h]
    \centering
    \begin{subfigure}{0.49\textwidth}
        \centering
        \includegraphics[width=\textwidth]{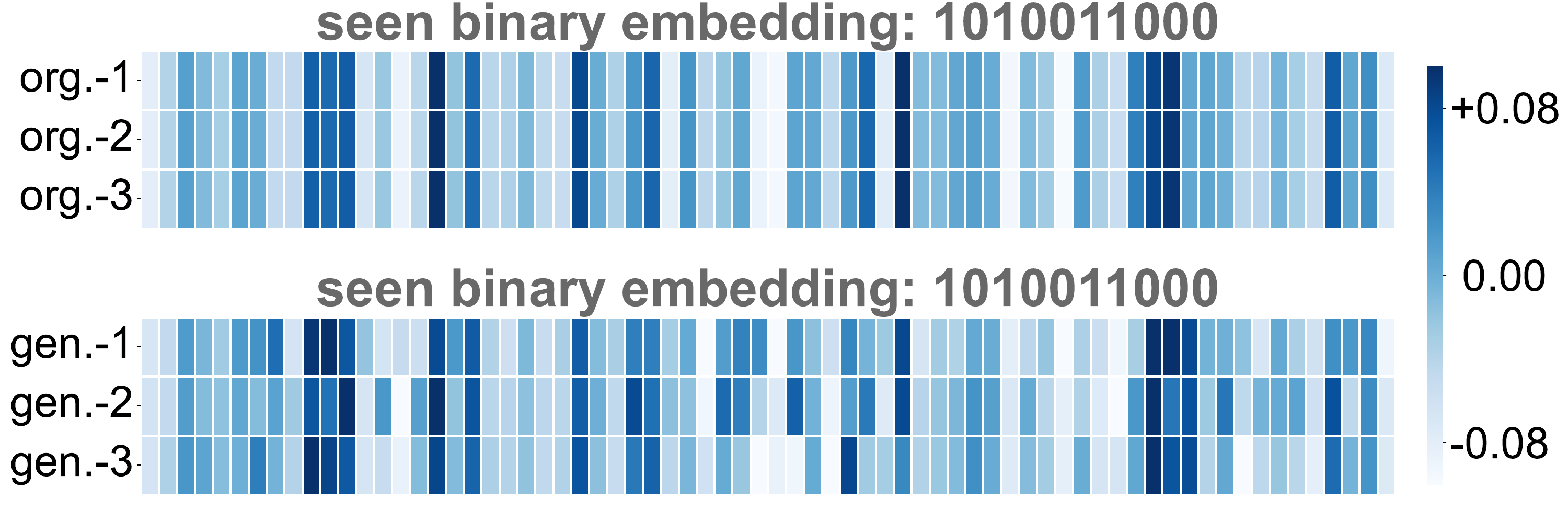}
        \caption{
        Original and generated models with identical seen binary embeddings are compared. 
        The three original models exhibit homogeneity,
while the generated models display diversity and maintain high
accuracy across all three models.}
        \label{fig:heatmap1}
    \end{subfigure}
    \hfill
    \begin{subfigure}{0.49\textwidth}
        \centering
        \includegraphics[width=\textwidth]{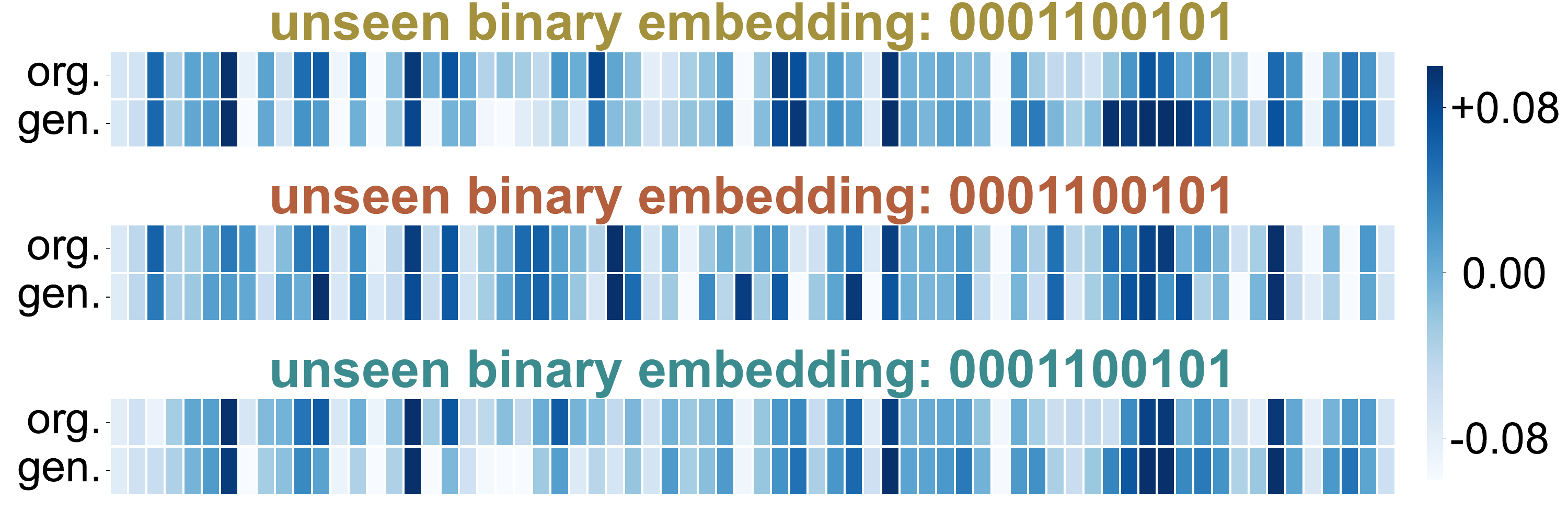}
        \caption{
         Original and generated models with 3 unseen binary embeddings
are compared. The results confirm that our approach can learn high-performing parameter patterns even when they are not included in
the training set.}
        \label{fig:heatmap2}
    \end{subfigure}
\vspace{-0.5em}
\caption{Illustration of the parameters of original and generated models in seen and unseen embeddings.
We select 100 parameters of the classification head and visualize its normalized values.}
\label{fig:heatmap}
\end{figure*}

\paragraph{Visualizations of model parameters.} 
We visualize the original and generated models for both seen and unseen tasks in Fig.~\ref{fig:heatmap}. For seen tasks, our approach generates diverse models compared to the original ones. Surprisingly, as shown in Fig.~\ref{fig:heatmap2}, we find that our approach can learn unseen parameter patterns. This demonstrates the potential generalization ability of our method.

\section{Related Works}

In this section, we mainly introduce recurrent models and previous works of parameter generation.

\vspace{-3mm}
\paragraph{Recurrent models.} 
Recurrent neural networks (RNNs) were first proposed to process sequential data. To tackle the vanishing gradient problem in early RNNs, long short-term memory (LSTM) \citep{hochreiter1991untersuchungen, hochreiter1997long} was introduced. Recently, transformer-based models~\cite{vaswani2017attention} exhibits excellent potential in sequential data processing, due to their parallelized training and scalability, such as linear attentions~\cite{wang2020linformer, choromanski2020rethinking},  RWKV~\cite{peng2023rwkv}, Mamba~\cite{gu2023mamba, dao2024transformers}, and xLSTM~\cite{beck2024xlstm}. 

\vspace{-1em}
\paragraph{Parameter generation.}
The core idea of parameter generation is to learn the distribution of trained parameters. Stochastic neural networks~\cite{sompolinsky1988chaos,bottou1991stochastic, graves2011practical} and Bayesian neural networks~\cite{neal2012bayesian, kingma2013auto, gal2016dropout} model the priors or probability distributions over the parameters. However, these methods are limited by their generality to large-scale parameter generation.
HyperNetworks~\cite{ha2017hypernetworks}, \textit{i.e.} is proposed to generate various architectures' parameters for a larger network.
Smash~\cite{brock2018smash} extends the range of architectures via a memory read-writes scheme.
With the development of diffusion models, many works~\cite{peebles2022learning, chou2023diffusion, erkocc2023hyperdiffusion, wang2024neural, soro2024diffusion, lin2024unleash, li2024text, jin2024conditional, peebles2022learning} adopt diffusion models to generate neural network parameters.
Hyper-Representations ~\cite{schurholt2022hyper, schurholt2022hyperpretrain, schurholt2024towards} use an autoencoder to capture the latent distribution of trained models.
COND P-DIFF~\cite{jin2024conditional} and Tina~\cite{li2024text} introduce text-controlled parameter generation method. 
Unfortunately, the above methods have a common drawback: \textit{can not generate large-scale parameters, such as whole parameters of ResNet, ViT, ConvNeXt, or LoRA.} Therefore, our approach brings new inspiration to the field of parameter generation.

\vspace{-2mm}
\section{Discussion and Conclusion}
\vspace{-1mm}
Our approach demonstrates promising results in large-scale parameter generation across various vision, language and unseen tasks.
However, we acknowledge that achieving true `AI generating AI' remains a distant goal.
Firstly, while our method shows potential in generating models for unseen tasks, it currently faces limitations in generating parameters for novel model architectures. 
Secondly, our approach is constrained by modeling parameter relationships within a single task, potentially limiting its practical applicability.
More importantly, future work should focus on simultaneously modeling parameter relationships across diverse architectures and tasks. Such an approach could yield a more powerful and versatile parameter generator, potentially advancing us closer to the `AI generating AI' era.
We hope our approach will inspire and encourage future research in neural network parameter generation.

\paragraph{Acknowledgments.} We sincerely thank Zhiyuan Liang, Zhuang Liu, Gongfan Fang, Xuanlei Zhao, Zangwei Zheng, Ziheng Qin, and Tianlong Chen for valuable discussions and feedbacks. This research is supported by the National Research Foundation, Singapore under its AI Singapore Programme (AISG Award No: AISG2-PhD-2021-08-008).

\bibliography{main}
\bibliographystyle{icml2025}

\clearpage
\appendix

We organize our appendix as follows. 

\textbf{Discussion with More Related Works}

Section~\ref{app_sec:overview_comparisons}: Overview of comparisons.\newline
Section~\ref{app_sec:related_HyperRepresentations}: Discussion with Hyper-Representations.\newline
Section~\ref{app_sec:G.pt}: Details and limitations of G.pt.\newline
Section~\ref{app_sec:pdiff}: Details and limitations of p-diff.\newline
Section~\ref{app_sec:diffusion_model}: More related works.

\textbf{Experimental Setting and Other Details}

Section~\ref{app_sec:recipe}: Training recipe.\newline
Section~\ref{app_sec:datasets}: Description of datasets.\newline
Section~\ref{app_sec:diffusion}: Detailed structure of recurrent diffusion.

\textbf{Additional Experimental Results}

Section~\ref{app_sec:permutation_state}: Effectiveness of permutation state.\newline
Section~\ref{more_results_of_sec4}: More results of Section~\ref{sec:4}.\newline
Section~\ref{app_sec:train_memory}: Training memory cost analysis.\newline
Section~\ref{app_sec:time}: Inference memory cost \& sampling time.\newline
Section~\ref{app_sec:sensiti}: Parameter sensitivity v.s. performance.\newline
Section~\ref{app_sec:data_collection}: Details of trained checkpoints collection.\newline
Section~\ref{app_sec:autoregression}: Why not auto-regression?

\section{Discussion with More Related Works} 
\label{app_sec:related}

\subsection{Overview of comparisons}
We compare RPG and four other methods from three aspects: scalability, performance, and generalization. Only our RPG excels in all three aspects simultaneously.

\label{app_sec:overview_comparisons}
\begin{table}[h]
    \centering
    \tablestyle{8pt}{1.2}
    \begin{tabular}{lccc}
    method &scalability &performance &generalization \\
    \shline
    $S_{\text{KDE30}}$%
    &\ding{56} &\ding{56} &\ding{56} \\
    p-diff%
    &\ding{56} &\ding{52} &\ding{56}\\
    SANE%
    &\ding{52} &\ding{56} &\ding{56} \\
    D2NWG%
    &\ding{52} &\ding{52} &\ding{56}\\
    RPG &\ding{52} &\ding{52} &\ding{52}\\
    \end{tabular}
    \vspace{-0.5em}
    \caption{Comparison of RPG and existing methods on key aspects: scalability, performance, and generalization.}
    \label{differences}
\end{table}

\subsection{Discussion with Hyper-Representations}
\label{app_sec:related_HyperRepresentations}
We mainly compare with three HyperRepresentation methods~\cite{schurholt2022hyper, schurholt2024towards, schurholt2022hyperpretrain}. These methods use an autoencoder to learn the latent features of trained models, so they call the latent feature HyperRepresentation.
This HyperRepresentation is then used for analyzing the model's performance or characteristics, or for sampling to generate new models or pre-trained parameters.
\begin{itemize}
\item  \cite{schurholt2022hyper} utilizes kernel density estimation (KDE) to sample model parameters on the learned HyperRepresentation space. They also emphasize the importance of layer-wise loss normalization in the learning process of HyperRepresentation. This work achieves parameter generation in small CNNs from Model Zoos~\cite{Schurholt2022modelzoos} with 2864 parameters.

\item \cite{schurholt2022hyperpretrain} focuses on using HyperRepresentation to sample the pre-trained model parameters. They also evaluate the ability of transfer learning by using a trained parameter autoencoder to initialize on unseen dataset. This work can be regarded as a cheap parameter initialization method.

\item \cite{schurholt2024towards} divides the neural network weights into subsets and utilizes a sequential autoencoder for neural embeddings (SANE) on there subsets with a sliding window. This work can generate the entire parameter of ResNet-18.
However, its generation process does not directly derive parameters from noise. Instead, it relies on a half-trained model for Kernel Density Estimation (KDE) sampling.
\end{itemize}

\vspace{0.3em}\noindent We summarize the main differences as follows:

\begin{itemize}
\item \textit{Hyper-representations as generative models} is hard to achieve comparable results as their original models that are used for training, but our approach obtains comparable results.
\item \textit{Hyper-representation for pre-training and transfer learning} focuses on parameter initialization while our approach targets to learn the distribution of high-performing neural network parameters.
\item \textit{SANE} is the latest method among these three HyperRepresentation methods. However, SANE uses a sliding window to model the relationship of a small part of trained parameters. Our approach uses a recurrent model among all parameters.

\item Our approach can synthesize many popular vision and language parameters, such as ConvNeXt-L and LoRA parameters of LLaMA-7B, with a maximum parameter count of approximately 200M, which is much larger than previous works.

Additionally, the parameters generated by our model can directly achieve almost peak performance without any finetuning. And the generation process is entirely synthesized from noise, eliminating the need for a few prompt examples that are trained for a few epochs, as required by the $S_{\text{KDE}}$~\cite{schurholt2024towards} sampling.

The capability of large-scale and high-accuracy generation makes our method more applicable in practical scenarios, significantly bridging the gap between theoretical parameter generation and practical application.

\end{itemize}

\subsection{Details and limitations of G.pt}
\label{app_sec:G.pt}
A primary limitation of G.pt~\cite{peebles2022learning} is the training data collection cost. By default, they collect 23 million checkpoints to train the parameter generator. Besides, they only evaluate the effectiveness of G.pt on small architectures, such as a low-dimensional MLP layer or a Convolutional layer with limited channels. The maximum number of generated parameters does not exceed 10,000.

\subsection{Details and limitations of p-diff}
\label{app_sec:pdiff}
P-diff \citep{wang2024neural} directly flattens all parameters into a 1D vector, disregarding the inter-layer parameter relationships. Furthermore, p-diff faces challenges in scaling up to large-scale parameter generation.

\subsection{More related works}
\label{app_sec:diffusion_model}
\paragraph{Diffusion models.}
Diffusion models~\cite{ho2020denoising, nichol2021improved, dhariwal2021diffusion} gain increasing popularity in recent years, due to their superiority in image generation. Ever since its advent, many works have been done focusing on improving the generation quality and efficiency of diffusion models.
For generation quality, \citet{rombach2022high} propose to conduct diffusion in the latent space, enabling high-resolution image synthesis. \citet{peebles2023scalable} leverage the transformer~\cite{vaswani2017attention} to explore scalability of diffusion models, proving the possibility of generating higher quality images by increasing model size. As for efficiency problem, 
efficient samplers~\cite{song2020denoising, lu2022dpm, song2023consistency}, efficiency models~\cite{fang2023structure, so2024temporal, yang2023diffusion}, and global acceleration approaches~\cite{ma2024deepcache, pan2024t} are proposed to increase diffusion models' efficiency. These methods facilitate generating high quality images with less computational and/or memory cost. Although diffusion models for image generation have achieved great success, improving quality and efficiency in large-scale parameter generation remains to be explored.

\section{Experimental Setting and Other Details}

\subsection{Training recipe} \label{app_sec:recipe}
In this section, we provide detailed training recipes and supplementary information. The number of parameters generated by our approach ranges from approximately 3K to 200M. The significant disparity necessitates different training settings. Generally, as the number of parameters increases, the learning process becomes more challenging, requiring higher training costs, particularly for generating parameters beyond 50 million. 
Therefore, our training settings are divided into two categories: the default setting and the setting for parameters over 50 million, shown in Tab. \ref{tab:training settings}.

\textbf{Data parallelism:} When the number of parameters is less than 50 million, we adopt a single GPU to run the training process. For larger number of parameters, we employ distributed data parallelism to facilitate the training.

\textbf{Diffusion batch size:} In our approach, the diffusion model is shared across all tokens. Typically, all tokens can be fed as a single batch into the diffusion model for training. However, in practice, we randomly select a subset of tokens from a long sequence for training, rather than feeding all parts at once. This approach significantly reduces memory usage without compromising performance. The `diffusion batch size' in Tab. \ref{tab:training settings} refers to the number of tokens fed into the diffusion model during a single training iteration.

\begin{table}[h]
    \centering
    \tablestyle{5pt}{1.2}
    \begin{tabular}{lcc}
    training setting & \#params. $<$ 50M & \#params. $>$ 50M \\
    \shline
    batch size & 16 & 8 \\
    optimizer & AdamW & AdamW \\
    learning rate & 3e-5 & 1e-5 \\
    training steps & 80,000 & 120,000 \\
    weight decay & 1e-5 & 1e-5 \\
    mixed precision & bfloat16 & bfloat16 \\
    diffusino batch size & 1024 & 512 \\
    \end{tabular}
    \vspace{-0.5em}
    \caption{Training recipe in detail. The amount of parameters that need to be generated is denoted as \textit{\#params.}}
    \label{tab:training settings}
\end{table}

\subsection{Description of datasets} \label{app_sec:datasets}
In this section, we introduce the datasets used in the paper, including those for classification, semantic segmentation, object detection\&instance segmentation, and commonsense reasoning.

\textbf{Classification:} 
\textbf{ImageNet-1k}~\cite{deng2009imagenet} is a large-scale visual database for visual object recognition research. It contains over 1 million images across 1000 categories and is widely used for training and benchmarking deep learning models. \textbf{CIFAR-10}~\cite{krizhevsky2009learning} dataset consists of 60,000 32$\times$32 colorful images in 10 different classes. It is commonly used for training machine learning and computer vision algorithms, providing a standard benchmark for image classification task.

\textbf{Semantic segmentation:} 
\textbf{ADE20K}~\cite{zhou2017scene} is a dataset for semantic segmentation and scene parsing, containing over 20,000 images annotated with pixel-level labels for 150 object categories. It is used to train models to understand and segment various objects and scenes in an image, making it valuable for applications in autonomous driving, robotics, and image editing.

\textbf{Instance segmentation \& object detection:} 
\textbf{COCO}~\cite{lin2014microsoft} dataset is a large-scale object detection, segmentation, and captioning dataset. It contains over 330,000 images in various resolutions, with more than 200,000 labeled instances across 80 object categories. COCO is widely used for training and evaluating models in object detection, segmentation, and image captioning tasks.

\textbf{Commonsense reasoning:} 
\textbf{BoolQ}~\cite{clark2019boolq}: Yes/No questions based on natural passages. \textbf{PIQA}~\cite{bisk2020piqa}: Questions about physical tasks and actions. \textbf{SIQA}~\cite{sap2019socialiqa}: Questions about social interactions. \textbf{HellaSwag}~\cite{zellers2019hellaswag}: Choosing the correct ending for stories. \textbf{ARC}~\cite{clark2018think}: Multiple-choice science questions. \textbf{OBQA}~\cite{OpenBookQA2018}: Questions  requires multi-step reasoning, commonsense knowledge, and rich text comprehension.

\subsection{Detailed structure of recurrent diffusion} \label{app_sec:diffusion}
In this section, we provide specific details about the proposed recurrent model and diffusion model in RPG. More detailed configurations can be found in Tab. \ref{tab:recurrent diffusion structure}.

\begin{table*}[htbp]
\centering
\small
    \tablestyle{12.0pt}{1.35}
    \begin{tabular}{c|c|cccc}
\shline%
    module & setting & RPG-Tiny & RPG-Small & RPG-Base & RPG-Large \\
\hline%
    \multicolumn{2}{c|}{adequate number of parameters} & $<$50K & 50K$\sim$10M & 5M$\sim$50M & $>$50M \\
\hline%
    \multirow{6}{*}{\makecell[c]{recurrent\\(Mamba)}} & d\_model of 1st layer & 256 & 4096 & 8192 & 12288 \\
    & d\_model of 2nd layer & 256 & 4096 & 8192 & 16384 \\
    & d\_state & 32 & 128 & 128 & 128 \\
    & d\_conv & 4 & 4 & 4 & 4 \\
    & expand & 2 & 2 & 2 & 2 \\
    & parameter counts & 1.3M & 256M & 1018M & 3076M \\
\hline%
    \multirow{10}{*}{\makecell[c]{diffusion\\(1D CNN)}} & encoder channels & (1, 32, 64, 128) & (1, 32, 64, 128) & (1, 32, 64, 128) & (1, 64, 96) \\
    & decoder channels & (128, 64, 32, 1) & (128, 64, 32, 1) & (128, 64, 32, 1) & (96, 64, 1)\\
    & token size & 256 & 4096 & 8192 & 16384 \\
    & kernel size & 7 & 7 & 7 & 7 \\
    & default solver & DDPM & DDPM & DDPM & DDIM \\
    & sampling steps & 1000 & 1000 & 1000 & 60 \\
    & $\beta$-start $\&$ $\beta$-end &(0.0001, 0.02) & (0.0001, 0.02) & (0.0001, 0.02) & (0.0001, 0.02) \\
    & betas schedule & linear & linear & linear & linear \\
    & number time steps & 1000 & 1000 & 1000 & 1000 \\
    & parameter counts & 0.3M & 17M & 69M & 273M \\
\shline%
    \end{tabular}
    \caption{Detailed information about four different sizes of recurrent diffusion. The \textit{adequate number of parameters} implies that our model is usually adequate to generate parameters in that scale, which is empirical results instead of an exact rule. It also necessitates considering other factors such as parameter sensitivity.}
    \label{tab:recurrent diffusion structure}
\end{table*}

\paragraph{Details of recurrent model.}
By default, the recurrent model consists of two Mamba layers~\cite{gu2023mamba}. As the increasing of parameters to generate, we need a larger recurrent model to capture the information in these parameters. The size of the recurrent model is mainly determined by the token size, which varies according to the number of parameters to be generated. Based on the token size, we categorize our model into four versions: Tiny, Small, Base, and Large. 

\paragraph{Details of diffusion model.}
Following p-diff~\cite{wang2024neural}, our diffusion model adopts a 1D convolutional architecture. The parameters of the diffusion model are significantly fewer than those of the recurrent model. We feed the prototypes from the recurrent model as conditions into the diffusion model by directly adding them to the feature map.

\paragraph{The setting of our main experiments.}
In our main experiments, ViT-Base, ConvNeXt-Large, ADE20K, COCO, and DoRA rank 64 used the setting of RPG-Large. CNN (s) and CNN (m) used the setting of RPG-Tiny. And all other experiments used the setting of RPG-Base.

\section{Additional Experimental Results}

\subsection{Effectiveness of permutation state}
\label{app_sec:permutation_state}

\paragraph{The effect of permutation state.}
RPG incorporates a permutation state operation to address parameter symmetries, which become particularly pronounced when collecting checkpoints from multiple training runs. To evaluate this, we collect checkpoints from different numbers of training runs (1, 3, and 10) and compare the performance with and without permutation state. These results demonstrate that permutation state operation effectively addresses parameter symmetries and enables stable results even when incorporating checkpoints from multiple training runs.

\begin{table}[H]
    \centering
    \tablestyle{6pt}{1.2}
    \begin{tabular}{c|c c  c}
   collected runs & org. & w/o permu. state & w/ permu. state \\
    \shline
    1 & 88.2 & 88.0 & 88.1\\
    3 & 88.3 & fail & 88.1 \\
    10& 88.5 & fail & 88.2 \\

    \end{tabular}
    \caption{Permutation state effectively mitigates parameter symmetries when collecting checkpoints from different training runs.}
    \label{tab: permutation_state_effect}
\end{table}

\subsection{More results of Section~\ref{sec:4}}
\label{more_results_of_sec4}
\paragraph{Results of generating models for unseen tasks.}
In Section~\ref{sec:4}, we show the potential of our approach in generating models for unseen tasks. In this part, we provide more results.  First, we compare the performance of original and generated models using all unseen embeddings in Tab.~\ref{app_tab:unseen}. Results demonstrate that our approach consistently achieves good results in unseen tasks.

\paragraph{Efficiency comparison.} To evaluate efficiency on unseen tasks, we compare 3 approaches: i) training ViT-Tiny from scratch, ii) finetuning an ImageNet-pretrained ViT-Tiny, and iii) finetuning a RPG-initialized model. As shown in Tab.~\ref{efficiency_comparison}, RPG significantly accelerates training for the unseen task.
These results establish RPG initialization as an effective approach that significantly reduces training costs.

\begin{table}[H]
    \centering
    \tablestyle{4pt}{1.2}
    \begin{tabular}{cc c  c}
    epoch &from scratch &finetune & RPG initialization + finetune \\
    \shline
    0 & 50.0 &52.9 &94.4\\
    1 & 61.4 &90.3 &96.8\\
    5 &69.1 &96.3 &97.3 \\
    10 &74.9 &97.1 &97.5 \\
    50 &86.7 &97.7 & 97.8 \\
    \end{tabular}
    \caption{Accuracy (\%) comparison for different training strategies on an unseen task. RPG initialization demonstrates superior performance and faster convergence.}
    \label{efficiency_comparison}
\end{table}
\vspace{-0.3em}

\begin{figure*}[htp]
    \centering
    \begin{subfigure}{0.45\textwidth}
        \centering
        \includegraphics[width=\textwidth]{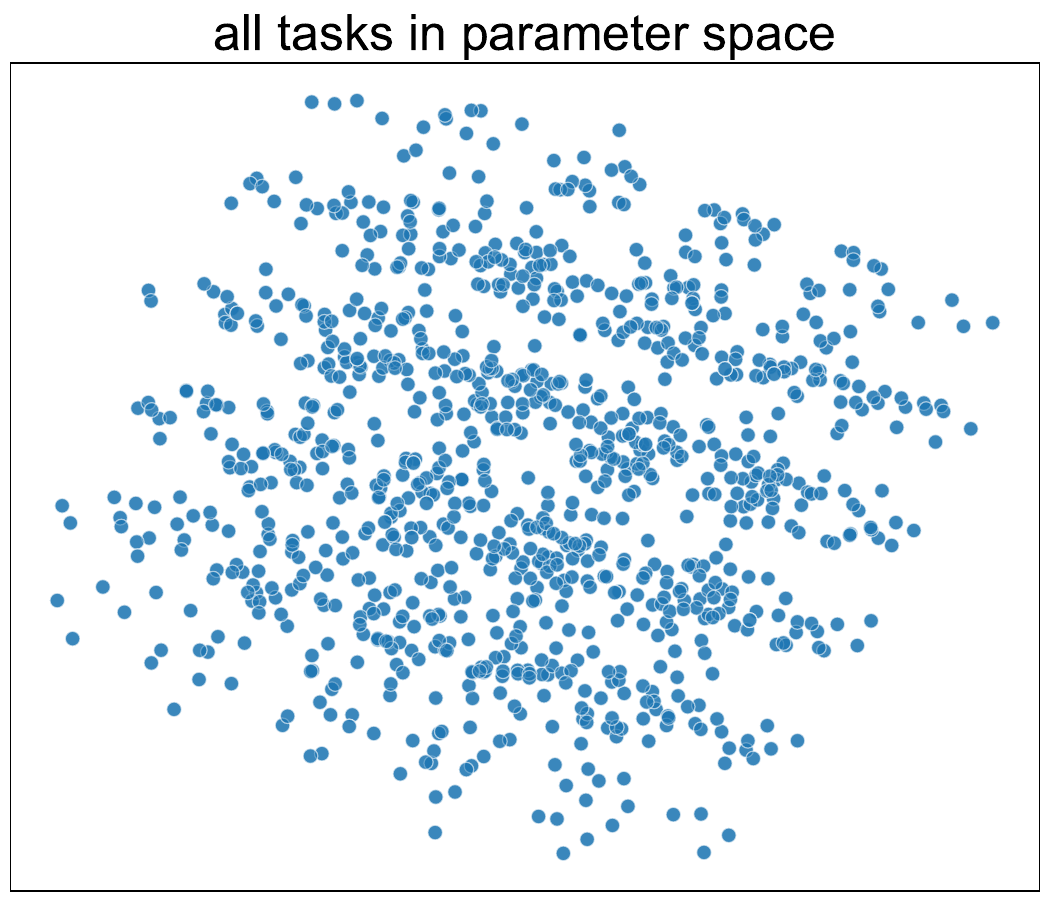}
        \caption{Visualization of the classification head of all 1022 tasks. This reveals that there is an inherent structure among the high-dimensional parameters, which can be captured by deep learning models.}
        \label{fig:all_tasks_parameter_space}
    \end{subfigure}
    \hspace{0.05\textwidth}
    \begin{subfigure}{0.45\textwidth}
        \centering
        \includegraphics[width=\textwidth]{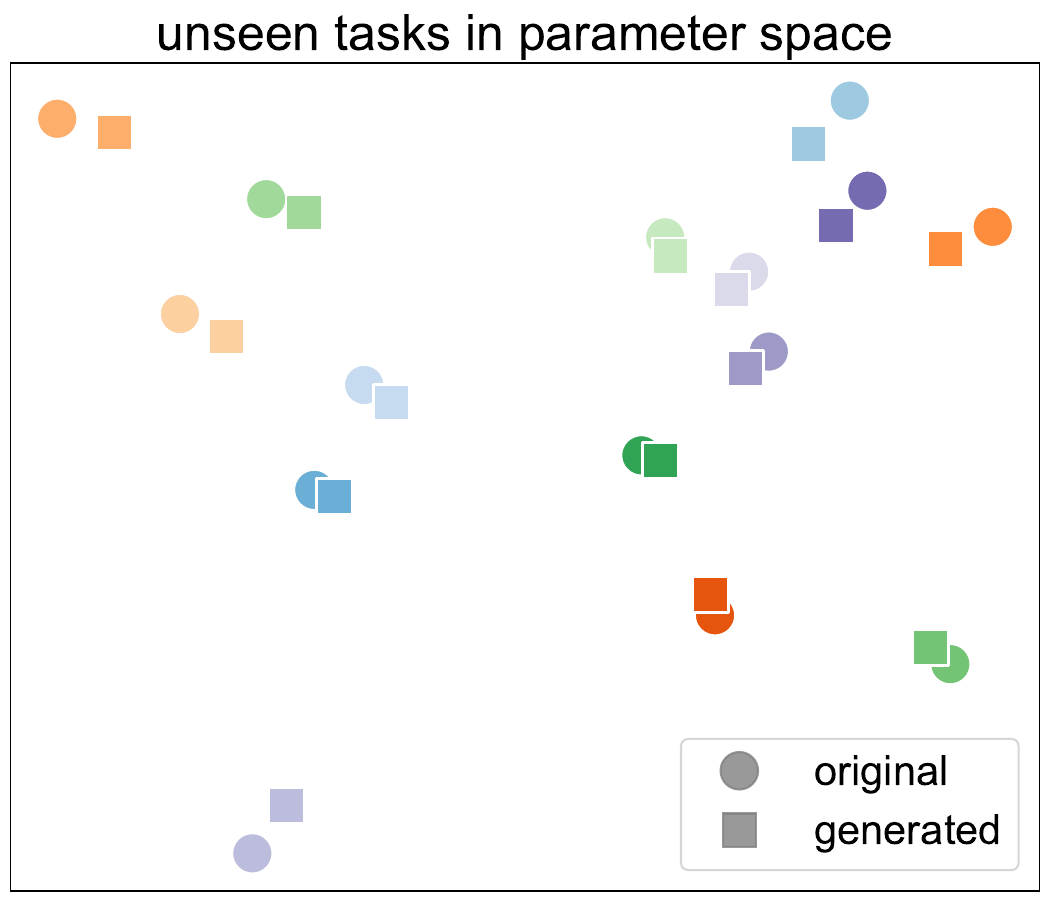}
        \caption{Visualization of the classification head in some unseen tasks. Parameters associated with the same task are indicated by a consistent color. Our method can capture the structure of the parameter space.}
        \label{fig:unseen_tasks_parameter_space}
    \end{subfigure}
    \caption{Principal Component Analysis (PCA) visualization of the classification head. The figures demonstrate the presence of an inherent structure in the parameter space and highlight our method's effectiveness in capturing this structure for unseen tasks.}
    \label{fig:parameter_space}
\end{figure*}

\paragraph{PCA visualization of classification head parameters.}
We also provide a visualization of the parameters of the classification head (a two-layer fully connected structure with total 38,976 parameters) for 1022 tasks as described in Section~\ref{sec:4} using Principal Component Analysis (PCA), which presents the structure of the parameter space in Fig.~\ref{fig:all_tasks_parameter_space}. Our generated model achieves an average accuracy of 91.2\% across all binary classification tasks, which indicates that our method has effectively learned this structure. 
Furthermore, we evaluate the parameters corresponding to unseen tasks and compared their positions in Fig.~\ref{fig:unseen_tasks_parameter_space} between the original and generated parameters. It is noteworthy that, even though the original parameters of these tasks are not included in the training data, the generated parameters consistently appeared in close proximity to the original ones. This observation further highlights the capability of our method to model the structure of the parameter space, even for tasks not previously encountered.

\begin{table*}[htp]
    \vspace{1em}
    \centering
    \tablestyle{12pt}{1.35}
    \begin{tabular}{@{}ccccc@{}}
        unseen binary embeddings & original & best acc. & average acc. & standard deviation \\
    \shline%
        \colorbox{code0!100}{0}\colorbox{code1!100}{1}\colorbox{code0!100}{0}\colorbox{code0!100}{0}\colorbox{code0!100}{0}\colorbox{code1!100}{1}\colorbox{code0!100}{0}\colorbox{code1!100}{1}\colorbox{code1!100}{1}\colorbox{code1!100}{1} & 97.3 & 94.4 & 93.9 & 0.6 \\
        \colorbox{code0!100}{0}\colorbox{code1!100}{1}\colorbox{code1!100}{1}\colorbox{code1!100}{1}\colorbox{code1!100}{1}\colorbox{code1!100}{1}\colorbox{code0!100}{0}\colorbox{code1!100}{1}\colorbox{code1!100}{1}\colorbox{code0!100}{0} & 98.1 & 96.6 & 94.9 & 2.1 \\
        \colorbox{code0!100}{0}\colorbox{code0!100}{0}\colorbox{code1!100}{1}\colorbox{code1!100}{1}\colorbox{code1!100}{1}\colorbox{code0!100}{0}\colorbox{code1!100}{1}\colorbox{code1!100}{1}\colorbox{code1!100}{1}\colorbox{code0!100}{0} & 97.4 & 95.0 & 94.2 & 1.1 \\
        \colorbox{code0!100}{0}\colorbox{code1!100}{1}\colorbox{code0!100}{0}\colorbox{code1!100}{1}\colorbox{code1!100}{1}\colorbox{code1!100}{1}\colorbox{code1!100}{1}\colorbox{code1!100}{1}\colorbox{code1!100}{1}\colorbox{code1!100}{1} & 98.4 & 96.1 & 95.8 & 0.3 \\
        \colorbox{code0!100}{0}\colorbox{code0!100}{0}\colorbox{code1!100}{1}\colorbox{code0!100}{0}\colorbox{code0!100}{0}\colorbox{code0!100}{0}\colorbox{code0!100}{0}\colorbox{code0!100}{0}\colorbox{code0!100}{0}\colorbox{code0!100}{0} & 98.9 & 96.6 & 95.2 & 2.3 \\
        \colorbox{code0!100}{0}\colorbox{code0!100}{0}\colorbox{code0!100}{0}\colorbox{code1!100}{1}\colorbox{code1!100}{1}\colorbox{code0!100}{0}\colorbox{code0!100}{0}\colorbox{code1!100}{1}\colorbox{code0!100}{0}\colorbox{code1!100}{1} & 96.7 & 92.9 & 91.6 & 1.1 \\
        \colorbox{code1!100}{1}\colorbox{code1!100}{1}\colorbox{code1!100}{1}\colorbox{code1!100}{1}\colorbox{code1!100}{1}\colorbox{code0!100}{0}\colorbox{code1!100}{1}\colorbox{code0!100}{0}\colorbox{code0!100}{0}\colorbox{code1!100}{1} & 97.6 & 94.8 & 94.1 & 0.7 \\
        \colorbox{code1!100}{1}\colorbox{code0!100}{0}\colorbox{code1!100}{1}\colorbox{code0!100}{0}\colorbox{code0!100}{0}\colorbox{code0!100}{0}\colorbox{code0!100}{0}\colorbox{code0!100}{0}\colorbox{code1!100}{1}\colorbox{code1!100}{1} & 98.1 & 95.7 & 91.8 & 3.7 \\
        \colorbox{code0!100}{0}\colorbox{code1!100}{1}\colorbox{code0!100}{0}\colorbox{code0!100}{0}\colorbox{code0!100}{0}\colorbox{code1!100}{1}\colorbox{code0!100}{0}\colorbox{code1!100}{1}\colorbox{code1!100}{1}\colorbox{code0!100}{0} & 97.1 & 93.6 & 90.7 & 4.3 \\
        \colorbox{code1!100}{1}\colorbox{code1!100}{1}\colorbox{code0!100}{0}\colorbox{code0!100}{0}\colorbox{code0!100}{0}\colorbox{code1!100}{1}\colorbox{code1!100}{1}\colorbox{code0!100}{0}\colorbox{code0!100}{0}\colorbox{code1!100}{1} & 97.0 & 94.0 & 90.1 & 3.6 \\
        
        \colorbox{code1!100}{1}\colorbox{code0!100}{0}\colorbox{code1!100}{1}\colorbox{code0!100}{0}\colorbox{code0!100}{0}\colorbox{code0!100}{0}\colorbox{code1!100}{1}\colorbox{code1!100}{1}\colorbox{code0!100}{0}\colorbox{code1!100}{1} & 97.3 & 91.3 & 90.7 & 0.8 \\
        \colorbox{code0!100}{0}\colorbox{code1!100}{1}\colorbox{code1!100}{1}\colorbox{code1!100}{1}\colorbox{code1!100}{1}\colorbox{code0!100}{0}\colorbox{code0!100}{0}\colorbox{code0!100}{0}\colorbox{code1!100}{1}\colorbox{code0!100}{0} & 96.3 & 95.4 & 89.4 & 6.3 \\
        \colorbox{code1!100}{1}\colorbox{code1!100}{1}\colorbox{code0!100}{0}\colorbox{code1!100}{1}\colorbox{code1!100}{1}\colorbox{code1!100}{1}\colorbox{code0!100}{0}\colorbox{code1!100}{1}\colorbox{code0!100}{0}\colorbox{code0!100}{0} & 97.6 & 92.6 & 90.5 & 3.2 \\
        \colorbox{code0!100}{0}\colorbox{code1!100}{1}\colorbox{code1!100}{1}\colorbox{code1!100}{1}\colorbox{code0!100}{0}\colorbox{code0!100}{0}\colorbox{code1!100}{1}\colorbox{code1!100}{1}\colorbox{code1!100}{1}\colorbox{code0!100}{0} & 96.3 & 90.8 & 89.1 & 1.9 \\
        \colorbox{code0!100}{0}\colorbox{code1!100}{1}\colorbox{code0!100}{0}\colorbox{code0!100}{0}\colorbox{code1!100}{1}\colorbox{code1!100}{1}\colorbox{code1!100}{1}\colorbox{code0!100}{0}\colorbox{code1!100}{1}\colorbox{code0!100}{0} & 96.3 & 91.9 & 88.4 & 4.4 \\
        \colorbox{code0!100}{0}\colorbox{code0!100}{0}\colorbox{code1!100}{1}\colorbox{code0!100}{0}\colorbox{code0!100}{0}\colorbox{code0!100}{0}\colorbox{code1!100}{1}\colorbox{code1!100}{1}\colorbox{code0!100}{0}\colorbox{code1!100}{1} & 97.5 & 93.7 & 88.0 & 5.6 \\
        \colorbox{code0!100}{0}\colorbox{code0!100}{0}\colorbox{code0!100}{0}\colorbox{code1!100}{1}\colorbox{code1!100}{1}\colorbox{code0!100}{0}\colorbox{code1!100}{1}\colorbox{code1!100}{1}\colorbox{code1!100}{1}\colorbox{code1!100}{1} & 96.5 & 90.8 & 85.5 & 7.0 \\
        \colorbox{code1!100}{1}\colorbox{code0!100}{0}\colorbox{code0!100}{0}\colorbox{code1!100}{1}\colorbox{code0!100}{0}\colorbox{code0!100}{0}\colorbox{code1!100}{1}\colorbox{code1!100}{1}\colorbox{code0!100}{0}\colorbox{code1!100}{1} & 96.4 & 86.7 & 83.7 & 3.6 \\
        \colorbox{code1!100}{1}\colorbox{code0!100}{0}\colorbox{code0!100}{0}\colorbox{code1!100}{1}\colorbox{code0!100}{0}\colorbox{code1!100}{1}\colorbox{code0!100}{0}\colorbox{code0!100}{0}\colorbox{code0!100}{0}\colorbox{code0!100}{0} & 97.7 & 85.6 & 83.2 & 2.0 \\
        \colorbox{code0!100}{0}\colorbox{code0!100}{0}\colorbox{code1!100}{1}\colorbox{code1!100}{1}\colorbox{code0!100}{0}\colorbox{code0!100}{0}\colorbox{code0!100}{0}\colorbox{code1!100}{1}\colorbox{code0!100}{0}\colorbox{code1!100}{1} & 96.3 & 90.2 & 79.2 & 9.6 \\

    \end{tabular}
    \captionof{table}{Performance comparisons between original and generated models on unseen tasks. Specifically, we generated 10 models for each unseen task, with binary embeddings in the unseen set as condition. The results consistently show that our generated models perform comparably to the original models.}
    \label{app_tab:unseen}
\end{table*}

\begin{figure*}[htp]
    \centering
    \includegraphics[width=0.9\textwidth]{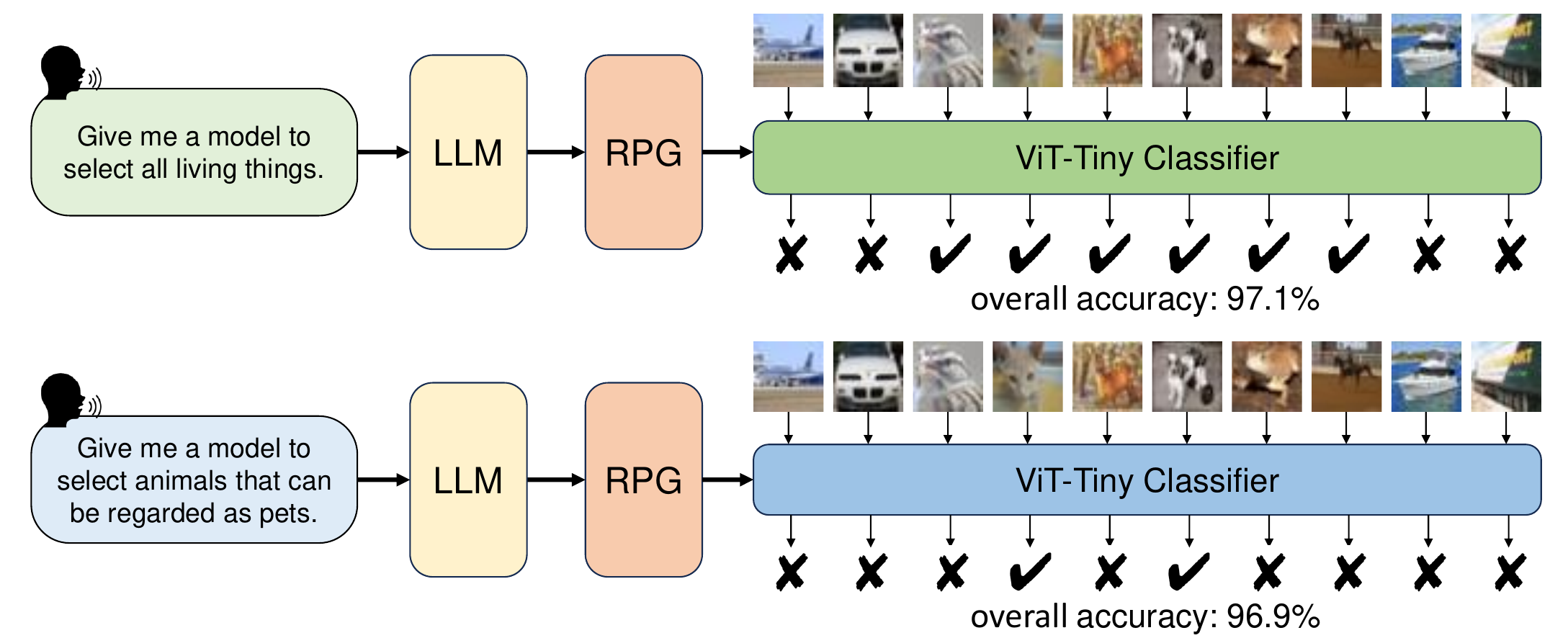}
    \caption{Illustration of RPG-generated models guided by binary embeddings from a large language model (Qwen2.5-3B~\cite{qwen2}), demonstrating neural network parameter generation conditioned by natural language.}
    \label{fig:condition_llm}
\end{figure*}

\begin{table*}[htp]
    \small
    \tablestyle{18pt}{1.2}
    \centering
    \begin{tabular}{>{\raggedright\arraybackslash}p{8cm} c c}
    \shline%
        prompt & expected embedding & acc. (\%) \\
    \hline%
        Give me a model to select all living things.
        & 0,0,1,1,1,1,1,1,0,0 & 98.7\\
        Find all vehicles that operate on roads.
        & 0,1,0,0,0,0,0,0,0,1 & 95.9\\
        Classify all mammals.
        & 0,0,0,1,1,1,0,1,0,0 & 97.6\\
        Select all man-made objects.
        & 1,1,0,0,0,0,0,0,1,1 & 97.7\\
        Find all things that are both man-made and can operate on water.
        & 0,0,0,0,0,0,0,0,1,0 & 98.4\\
        Select all animals that can be regarded as pets.
        & 0,0,0,1,0,1,0,0,0,0 & 96.8\\
        Select all things that can fly.
        & 1,0,1,0,0,0,0,0,0,0 & 87.7\\
        Find all animals with fur.
        & 0,0,1,1,1,1,0,1,0,0 & 70.4\\
        Select all pets commonly found in households.
        & 0,0,1,1,0,1,0,0,0,0 & 83.3\\
        Identify all cold-blooded animals.
        & 0,0,0,0,0,0,1,0,0,0 & 98.9\\
    \shline%
    \end{tabular}
    \caption{The table demonstrates examples of generating from abstract prompts using LLM and RPG. We are able to achieve high accuracy on abstract prompts. The \textit{expected embedding} is used for evaluating the accuracy.}
    \label{tab:example_llm_and_rpg}
\end{table*}

\begin{figure*}[htbp]
    \centering
    \begin{subfigure}{0.42\textwidth}
        \centering
        \includegraphics[width=\textwidth]
        {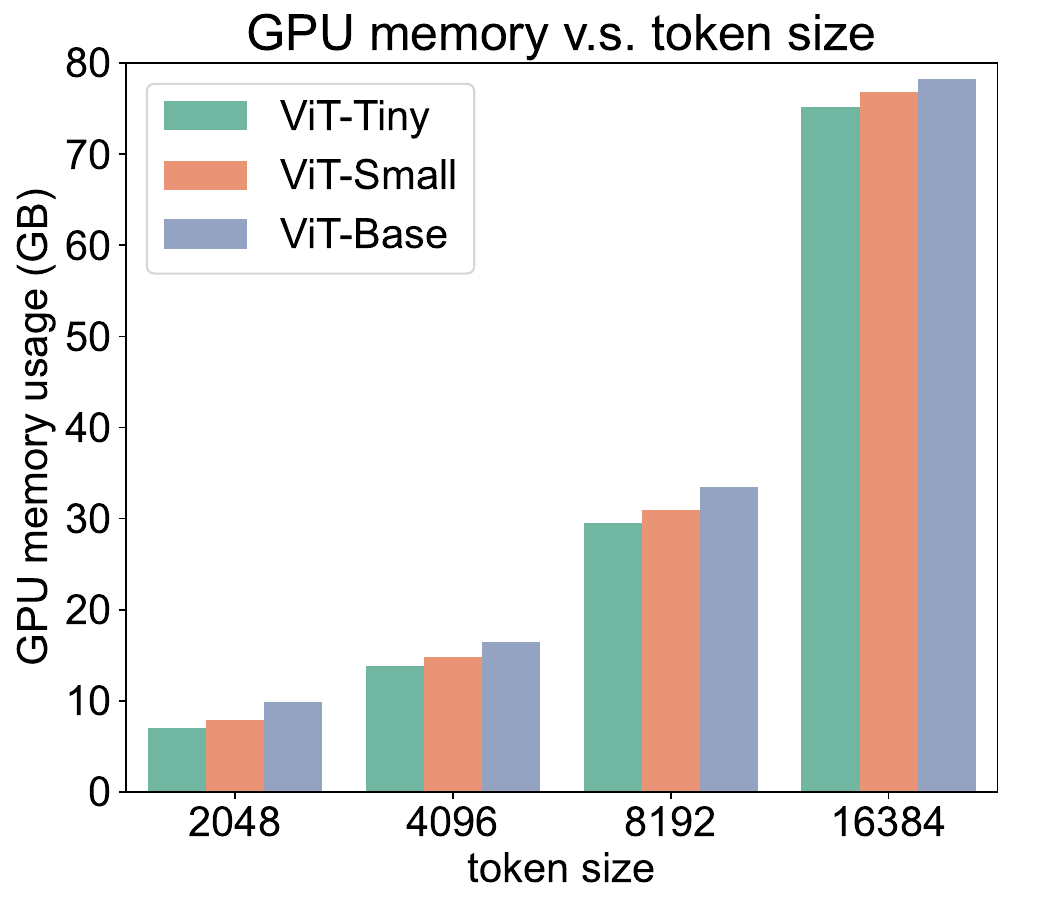}
        \caption{Visualization of GPU memory v.s. token size. GPU memory usage increases proportionally to the token size. Thus, the token size cannot get larger infinitely; we need to choose a proper token size.}
        \label{appendix_token_size_memory}
    \end{subfigure}
    \hspace{0.05\textwidth}
    \begin{subfigure}{0.42\textwidth}
        \centering
        \includegraphics[width=\textwidth]{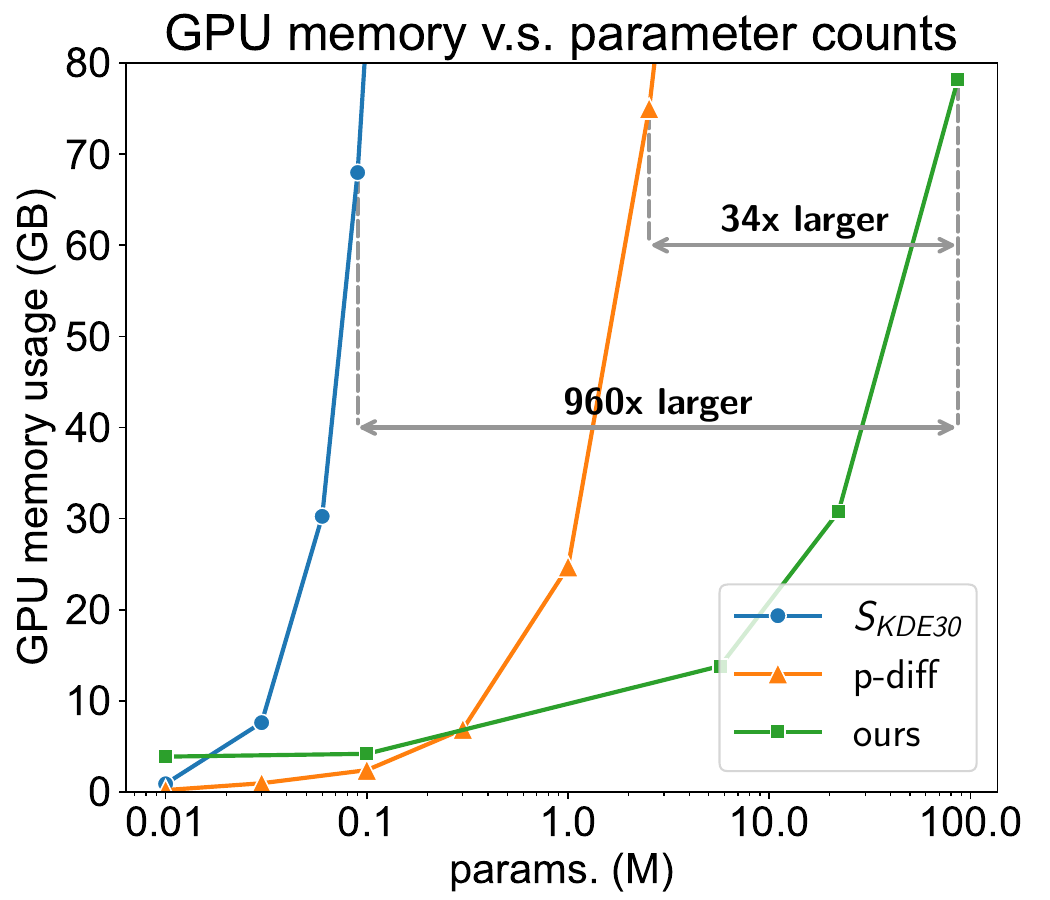}
        \caption{Visualization of GPU memory v.s. parameter counts. Our method can generate much more parameters than existing approaches e.g. $\textit{\text{S}}_{\textit{\text{KDE30}}}$~\cite{schurholt2022hyper} using a single NVIDIA H100 80G GPU.}
        \label{appendix_params_memory}
    \end{subfigure}
    \vspace{-0.5em}
    \caption{Training memory cost analysis. \textit{Left:} GPU memory v.s. token size. \textit{Rihgt:} GPU memory v.s. parameter counts.}
\end{figure*}

\paragraph{Conditional generation guided by LLM.} 
To demonstrate the application scenarios of our model, we utilize large language model to generate binary embeddings to guide RPG in generating corresponding classification models. For the first example in Fig~\ref{fig:condition_llm}, we give the LLM a prompt: `Give me a model to select all living things.' With the binary embedding provided by the LLM, our RPG then generates a ViT-Tiny classifier. After that, We use images in CIFAR-10 to evaluate the model's accuracy. The model should classify `bird', `cat', `deer', `dog', `frog', and `horse' to the positive class, which we used as the ground truth. The result is 97.1\%. Some other examples are in Tab~\ref{tab:example_llm_and_rpg}. This experiment demonstrates our method's capability to generate neural network parameters based on natural language guidance, highlighting the potential applications of our method.

\subsection{Training memory cost analysis}
\label{app_sec:train_memory}

In this section, we analyze the GPU memory utilization during training.
GPU memory consumption is usually highly correlated with two factors: i) the size of the generative model and ii) the size of generated parameters.
We analyzed the impact of these two factors on the GPU memory utilization during the training of our approach.

\paragraph{GPU memory v.s. token size}
We visualize the GPU memory usage with different token sizes in Fig.~\ref{appendix_token_size_memory}.
As the token size increases, the scale of the recurrent model significantly grows, leading to a notable increase in GPU memory consumption. This implies that, when the performance of the generated models is comparable, we prefer to use models with smaller token sizes.

\paragraph{GPU memory v.s. parameter counts}
We conduct experiments to show the relationship between GPU memory and generated parameter counts in Fig.~\ref{appendix_params_memory}.
In previous methods, the relationship between GPU memory consumption and the number of parameters in the generated model was quadratic~\cite{schurholt2022hyper} or directly proportional~\cite{wang2024neural}.
This limits their practicality and application range.
In contrast, our approach demonstrates remarkable efficiency: with equivalent GPU memory usage, it can generate models with 34 to 960 times more parameters compared to previous methods.

\subsection{Inference memory cost \& sampling time} \label{app_sec:time}

\begin{table*}[htp]
\centering
\small
    \tablestyle{9pt}{1.2}
    \begin{tabular}{c|c|ccccc}
    metrics & inference mode & ResNet-18 & ResNet-50 & ViT-Tiny & ViT-Small & ConvNeXt-A \\
    \shline%
    \multirow{3}{*}{time (minute)} 
    & sequential & 18.6 & 38.0 & 9.8 & 33.8 & 6.8 \\
    & partially parallel & 1.8 & 3.3 & 1.1 & 2.9 & 0.9 \\
    & fully parallel & 1.7 & 3.3 & 1.1 & 2.9 & 0.9 \\
    \hline%
    \multirow{3}{*}{memory cost (GB)} 
    & sequential & 6.3 & 6.4 & 6.2 & 6.4 & 6.2 \\
    & partially parallel & 10.3 & 10.5 & 10.3 & 10.5 & 10.3 \\
    & fully parallel & 30.8 & 50.5 & 19.4 & 45.9 & 15.2 \\
    \end{tabular}
    \caption{We show the inference time and memory cost under different inference modes. All information in this table is collected from a single NVIDIA H100 80G GPU. We report the time and memory required to generate a single model.}
    \label{extra_result_steps}
\end{table*}

\begin{table*}[htp]
\centering
\begin{center}
\tablestyle{10pt}{1.2}
    \begin{tabular}{c| cc|cccc}
    \multirow{2}{*}{model} & \multirow{2}{*}{params. (M)} & \multirow{2}{*}{sensitivity} & \multicolumn{4}{c}{accuracy decline}  \\
    &  &  & ours & noise (0.01) & noise (0.10) & noise (1.00) \\
\shline
    ConvNeXt-A & 3.7   & +++ & 0.85 & 62.83 & 0.60 & 0.03 \\
    ResNet-18 & 11.7 & ++ & 0.39  & 53.56  & 0.46  & 0.00 \\
    ViT-Base & 86.6  & + & 0.09 & 5.39 & 0.02  & 0.00 \\
    \end{tabular}
\end{center}
\vspace{-0.5em}
\caption{The accuracy decline reflects the accuracy gap between the original model and the generated model or the model after adding noise.  We add Gaussian noise with weights of 0.01, 0.10, and 1.00 to the parameters to measure model sensitivity. Results demonstrate that smaller models are relatively more sensitive than larger ones. The more plus signs (+) , the higher the sensitivity.}
\label{app_tab:sentivity}
\end{table*}

In this section, we present more information about the sampling, including memory usage, inference time, and the balance between sequential and parallel inference.
In Tab. \ref{tab: result_steps}, we show the sampling time and memory usage for ViT-Base and ConvNeXt-L. Here, we present the sampling time and memory usage for other models. In Tab. \ref{extra_result_steps}, we adopt DDPM as the solver and conduct 1000-step sampling. Since the diffusion model in RPG is shared among all the parameter tokens, we can adopt different inference modes to find a balance between memory usage and inference speed: 
\begin{itemize}
    \vspace{0.2em} \item \textbf{fully parallel:} All tokens are fed into the diffusion model simultaneously. This approach results in a high memory usage but  achieves a high generation speed.
    \vspace{0.2em} \item \textbf{sequential:} Tokens are fed into the diffusion model one by one. This approach significantly reduces memory usage, as the model only occupies memory for inferring a single token at a time. This enable us to generate parameters of models listed on a GPU with less than 8GB of memory .
    \vspace{0.2em} \item \textbf{partially parallel (default):} In partial parallel mode, we set 256 tokens as a batch for the diffusion model inference. This approach significantly boosts speed with a slight increase in GPU memory usage, reaching an optimal trade-off between memory and speed. We adopt this as the default setting.
\end{itemize}
\vspace{0.5em}\noindent Based on the results in Tab.~\ref{extra_result_steps}, our approach can be flexibly applied to many other GPUs as it can achieve a good trade-off between memory and time.

\subsection{Parameter sensitivity v.s. performance}

\label{app_sec:sensiti}
According to conventional understanding, larger parameter quantities are generally more challenging to learn. However, our experiments reveal that this rule is not absolute and demonstrates instability in learning some small model parameters.  

This motivates us to investigate the relationship between parameter sensitivity and generation quality. Specifically, we add Gaussian noise with weights of 0.01, 0.10, and 1.00 to the original parameters to measure model sensitivity, as shown in Tab.~\ref{app_tab:sentivity}. We observe that as noise weight increases, performance decreases for all models, with smaller models being more affected than larger ones. This indicates that smaller models are relatively more sensitive. Additionally, we notice that the performance gap between the original and generated models widens as sensitivity of the model increases. This demonstrates a strong correlation between a model's sensitivity and the difficulty of generating its parameters.

\subsection{Details of trained checkpoint collection}
\label{app_sec:data_collection}
This section mainly supplements the collection of checkpoints in Section~\ref{sec:experiments}.
First, we obtain the pre-trained models, which either come from model libraries (such as timm~\cite{rw2019timm}) or are trained by ourselves. Then, we fine-tune the full model for one epoch, and save 50 checkpoints during this epoch uniformly. 
In Tab~\ref{tab: permutation_state_effect}, the term \textit{collected runs} refers to the number of times the entire process, from pre-training to fine-tuning and saving, is repeated. This is done without fixing the seed, resulting in checkpoints from entirely different permutations.

\subsection{Why not auto-regression?}
\label{app_sec:autoregression}

It is worth noting that our approach does not employ an auto-regressive method, \textit{i.e.}, we do not feed the output back as input again. Our method takes position embedding and permutation state as inputs and synthesizes neural network parameters as outputs, forming a standard recurrent neural network. We have attempted to train our model using an auto-regressive approach such as a decoder-only transformer architecture, whose results are shown in Tab~\ref{tab:autoregressive}. Due to the accumulation of errors during the auto-regressive process in inference, the parameters generated at the end of sequence become nearly indistinguishable from noise, leading to poor performance. In contrast, in RPG, noise is only introduced in the diffusion model and does not accumulate in the recurrent process, ensuring stable parameter generation.

\begin{table}[H]
    \centering
    \tablestyle{10pt}{1.2}
    \begin{tabular}{lccc}
    & best & avg. & min. \\
    \shline
    original & 75.2 & 74.9 & 74.7 \\
    auto-regressive & fail & fail & fail \\
    non-auto-regressive & 75.0 & 74.8 & 74.6 \\
    \end{tabular}
    \vspace{-0.5em}
    \caption{Comparison between auto-regressive and non-auto-regressive methods. The auto-regressive model refers to a transformer decoder-only structure, while the non-auto-regressive model refers to a transformer encoder-only structure with the same scale. To ensure a fair comparison, we also applied a causal mask to the transformer encoder-only structure, ensuring that information can only be passed in one direction.}
    \label{tab:autoregressive}
\end{table}

\end{document}